\documentclass[lettersize,journal]{IEEEtran}
\usepackage{amsmath,amsfonts}
\usepackage{algorithmic}
\usepackage{algorithm}
\usepackage{array}
\usepackage[caption=false,font=normalsize,labelfont=sf,textfont=sf]{subfig}
\usepackage{tcolorbox} 
\usepackage{amsmath}
\usepackage{textcomp}
\usepackage{stfloats}

\usepackage{url}
\usepackage{verbatim}
\usepackage{graphicx}
\usepackage{subcaption}
\usepackage{booktabs}
\usepackage{colortbl}
\usepackage{color, xcolor} 
\usepackage{enumitem}
\usepackage{cite}

\begin{document}

\title{Consistency-Aware Editing for Entity-level Unlearning in Language Models}

\author{Xiaoqi Han, 
V\'ictor Guti\'errez-Basulto, 
Ru Li, 
Xiaoli Li, 
Jiye Liang, 
Jeff Z.Pan 
\thanks{Xiaoqi Han, Jiye Liang and Ru Li are with the Shanxi University, China}
\thanks{V\'ictor Guti\'errez-Basulto is with the Cardiff University, UK}
\thanks{Xiaoli Li is with the Singapore University of Technology and Design, Singapore}
\thanks{Jeff Z. Pan is with ILCC, School of Informatics, University of Edinburgh, Edinburgh, UK}
\thanks{The corresponding authors are Ru Li, V\'ictor Guti\'errez-Basulto and Jeff Z.Pan}
}
% The paper headers
%\markboth{Journal of \LaTeX\ Class Files,~Vol.~14, No.~8, August~2021}%
%{Shell \MakeLowercase{\textit{et al.}}: A Sample Article Using IEEEtran.cls for IEEE Journals}

% \IEEEpubid{0000--0000/00\$00.00~\copyright~2021 IEEE}
% Remember, if you use this you must call \IEEEpubidadjcol in the second
% column for its text to clear the IEEEpubid mark.

\maketitle

\begin{abstract}
Large language models (LLMs) risk retaining sensitive, copyrighted, or harmful information from their training data.
Entity-level unlearning addresses this issue by removing all knowledge of a specific entity while preserving the model's overall capabilities. 
Existing approaches typically rely on full-model fine-tuning or prompt-based interventions, which can be computationally expensive or brittle when handling paraphrased queries.
Recently, model editing has emerged as an efficient alternative for updating knowledge in LLMs, offering a promising direction for unlearning.
However, existing editing techniques are typically designed for instance-level updates, modifying responses to specific attributes of an entity rather than eliminating all knowledge associated with the entity.
In this paper, we investigate how editing techniques can be adapted for effective and efficient entity-level unlearning.
To this end, we introduce a novel \textbf{consistency-aware editing (CAE)} framework.
CAE aggregates a diverse set of prompts related to a target entity, including its attributes, relations, and adversarial paraphrases.
It then jointly learns a low-rank update guided by a consistency regularizer that aligns the editing directions across prompts.
This promotes robust and comprehensive forgetting while minimizing interference with unrelated knowledge.
We further examine where different entities are stored within the model and how many diverse prompts are needed for successful unlearning.
We evaluate CAE on two challenging benchmarks, RWKU and ToFU, and demonstrate that it
(i) provides insights into how entity-level knowledge is internally represented and deleted in LLMs,
(ii) significantly improves forgetting accuracy and robustness over traditional unlearning and editing baselines, and
(iii) enables scalable entity removal using only tens of carefully selected prompts. %Code is available at \url{https://anonymous.4open.science/r/CAEditor-A8B1/}.
\end{abstract}

\begin{IEEEkeywords}
Knowledge Unlearning, Model Editing, Large Language Models
\end{IEEEkeywords}

\section{Introduction}

Large language models (LLMs) \cite{touvron2023Llama,chatglm,openai2024gpt4} have exhibited remarkable performance across a broad spectrum of tasks. 
Nevertheless, they frequently and inadvertently memorize undesirable content, such as copyrighted material, sensitive personal data, and toxic or biased language \cite{pan2020privacy, karamolegkou2023copyright}.
Such unintended retention introduces serious privacy, security, and ethical risks,
%The inclusion of such content raises serious privacy, security, and ethical concerns, 
thereby constraining the safe and responsible deployment of LLMs in real-world settings.
To mitigate these risks,  \emph{knowledge unlearning} \cite{bourtoule2021machine, jang2023knowledge,si2023knowledge,liu2025rethinking} has been proposed as a promising approach to erase specific unwanted knowledge from the model while preserving its overall capabilities.
Existing knowledge unlearning methods predominantly rely on gradient ascent fine-tuning over the data intended for removal \cite{eldan2023s, mainitofu, yao2024large, liu2024towards}, or on steering model outputs via representation engineering and in-context prompting \cite{li2024wmdp, liu2024large, pawelczyk2024context}.
While these approaches can be effective, they primarily address instance-level unlearning, which targets the deletion of specific facts or expressions drawn from a curated forget set.
However, they often fall short in the context of entity-level unlearning, which aims to comprehensively eliminate information related to an entire entity \cite{ma2025unveiling, chang2025retain}.
Unlike instance-level methods that fine-tune the model on discrete snippets of knowledge, entity-level unlearning seeks to erase all associated knowledge about a subject, regardless of phrasing or context.
Current techniques 
%often fail to achieve this depth of removal, as they overfit to specific formulations and struggle to generalize across the diverse ways an entity can be described.
tend to overfit to the surface form of training prompts and struggle to generalize across the diverse linguistic realizations of an entity.
Furthermore, expanding the forget set to improve coverage frequently results in substantial collateral damage, diminishing the model's performance on unrelated facts or closely related entities \cite{ma2025unveiling, chang2025retain}.

Recently, model editing \cite{ke, mend, rome} has emerged as a powerful technique for modifying factual knowledge in LLMs.
Building on this foundation, we frame knowledge unlearning as a special case of model editing: %in this case the goal is not to rewrite a model's output to a specific fact, but to induce uncertainty 
rather than rewriting a model’s output to a specific fact, the objective is to induce uncertainty
(e.g., responding with “I don't know''), thereby removing undesired knowledge from the model.
While prior work has applied location-based editing methods to the unlearning setting with encouraging results \cite{li2025editing, wang2024large, nvestigating_Model_Editing}, these efforts have been largely focused on instance-level knowledge.
A key open challenge remains: \emph{how to extend location-based editing methods to support entity-level unlearning effectively?}

\begin{figure}[htbp]
    \centering
    \includegraphics[width=1\linewidth]{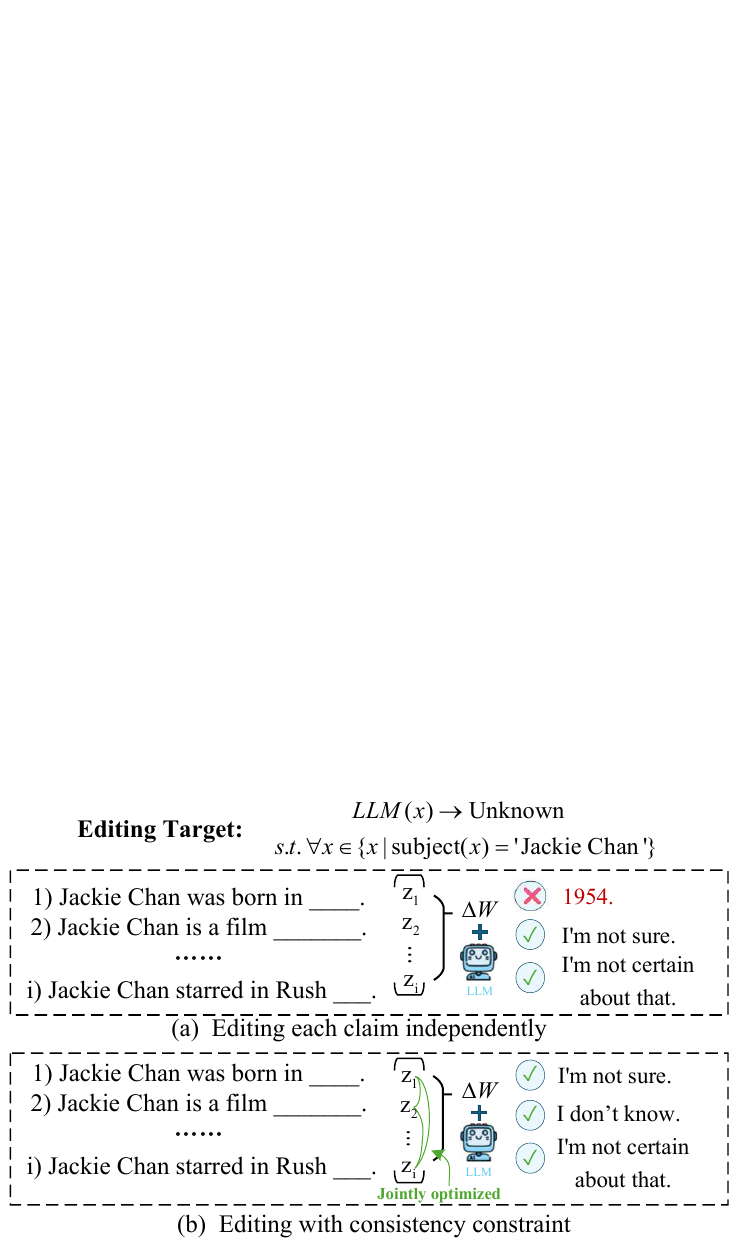}
    \caption{
    We compare two editing strategies on unlearning the entity of {Jackie Chan}. 
    (a) Editing each prompt independently leads to inconsistent editing directions ($z_i$) and partial forgetting. 
    (b) Joint optimization with a consistency constraint aligns $z_i$ vectors, resulting in a shared update $\Delta W$ that generalizes across all facts. 
    }
    \label{motv}
\end{figure}

In this paper, we aim to understand how language models store facts about the same entity, and why existing location-based editing methods fall short in the entity-level unlearning.
We observe that although these methods can effectively modify specific facts, they typically treat each input independently. 
As illustrated in Figure~\ref{motv} (a), location-based editing methods typically calculate  %parameter shifts 
a parameter shift  ($\Delta W$) for each input fact ($z_*$) independently. 
However, this independent optimization often results in inconsistent edits, where updates for different facts may conflict or only partially erase the target knowledge. 
This is especially problematic when the entity has many interrelated properties stored across the model in diverse forms, making isolated edits insufficient for complete forgetting.

To address this, we propose the \textbf{C}onsistency‑\textbf{A}ware \textbf{E}diting (CAE) framework for entity-level unlearning in LLMs.
We start by retrieving relevant facts about a target entity from Wikidata\footnote{\url{https://query.wikidata.org}}, and design an SVD‑based selection strategy to identify the most valuable facts for a entity.
For each valuable fact, CAE derives a separate edit vector, and introduces a consistency constraint that minimizes variance across these vectors, encouraging them to align in a common direction. 
This joint optimization (Figure~\ref{motv}(b)) results in a coherent parameter update that consistently suppresses all factual expressions of an entity, enabling robust and unified unlearning.
To evaluate the effectiveness of CAE at the entity-level, we conduct extensive experiments on two benchmarks: RWKU \cite{jinrwku} and ToFU \cite{mainitofu}. 
We compare CAE against a range of unlearning algorithms and model editing methods, and show that CAE achieves superior forgetting performance and efficiency.
In summary, our contributions are as follows: %\nb{V: I will read the contributions after they are updated}

(1) We identify where entity-level knowledge is stored in LLMs and reveal %show 
the limitations of existing location-based editing methods for %in 
entity-level unlearning, particularly the inconsistency introduced by independently editing multiple prompts associated with the same entity. %related to an entity.

(2) We introduce an SVD-based fact selection strategy and propose Consistency-Aware Editing (CAE), a novel editing-based unlearning approach that jointly optimizes the valuable facts under a consistency constraint, enabling %to achieve 
coherent and comprehensive entity-unlearning.

(3) We empirically validate CAE on the RWKU and ToFU benchmarks, showing that it consistently outperforms prior unlearning and editing methods in both effectiveness and efficiency.  Furthermore, CAE offers a more favorable and stable trade-off between forgetting and retention.

\section{Related Work}

\smallskip \noindent \textbf{Unlearning in Large Language Models }  

A widely adopted strategy is gradient-based unlearning, where the model is fine-tuned or supervised with counterfactual signals, training on a designated \emph{forget set} to reduce the likelihood of generating specific target facts \cite{eldan2023s, yao2024large, liu2024towards,zhang2025llm}. While effective at suppressing memorized knowledge, this approach is computationally intensive and depends heavily on carefully curated deletion examples.

An alternative direction avoids modifying model parameters and instead performs unlearning at inference time. These methods include %inserting negative exemplars or suppression instructions into the prompt 
augmenting prompts with negative exemplars or suppression instructions
\cite{liu2024large, pawelczyk2024context}, or introducing auxiliary triggers at the representation level to steer outputs away from target knowledge \cite{li2024wmdp}.

However, most existing approaches are evaluated under instance-level unlearning settings, where only isolated facts are removed. This setup does not reflect real-world scenarios, where the objective is often to forget all knowledge related to a specific entity or concept.
To bridge this gap, recent benchmarks such as ToFU \cite{mainitofu} and RWKU \cite{jinrwku} have been introduced, which focus on entity-level unlearning. These datasets organize related facts under common subjects, enabling a more comprehensive and entity-centric evaluation.
Entity-level unlearning thus poses greater challenges due to the interdependence and redundancy of entity-associated knowledge, requiring more robust and generalizable methods.

\smallskip \noindent \textbf{Model Editing }  
% Model editing offers a middle ground between full fine-tuning and inference-time interventions by introducing targeted weight updates that change specific knowledge in LLMs with minimal retraining. 
% These methods 
Model editing aim to localize and modify parameters associated with the target knowledge, while preserving the model's overall performance and avoiding widespread collateral effects.
Notable approaches include ROME \cite{rome}, which identifies and updates MLP subspaces with a rank-one weight change.
MEMIT \cite{memit} extends rank-1 edits to mass-editing thousands of facts.
MEND \cite{mend} trains a network to predict gradient modifications for one-shot edits.  
These techniques require only a handful of examples and %preserve most of 
largely preserve the model's original capabilities, making them highly attractive for rapid, localized corrections.
Recent work by \cite{wang2024large} and \cite{nvestigating_Model_Editing} has begun to investigate model editing as a means of enabling unlearning. However, these studies remain limited to instance-level unlearning. When applied independently to multiple prompts related to the same subject, existing editing methods often produce divergent update directions. This leads to inconsistent forgetting, where some paraphrases are effectively suppressed while others still elicit the original knowledge. Consequently, such methods fall short of achieving comprehensive, entity-level unlearning, which demands coordinated updates across semantically related facts.

\section{Preliminaries}
\label{Task}
\subsection{Entity-level Unlearning}
Entity-level unlearning aims to remove all knowledge associated with one or more entities from a trained model. Formally, let $\mathcal{M}(\cdot; \theta)$ denote a model trained on a dataset $D$ with parameters $\theta$. %, and l
Let $E_f = \{e_1, e_2, \ldots, e_m\}$ be a set of $m$ entities, where each entity $e_i$ is associated with a set of facts $\mathcal{K}(e_i) \subset D$. The complete forget set is then defined as $\mathcal{K}_f = \bigcup_{i=1}^m \mathcal{K}(e_i)$,
containing all facts that must be removed from the model.
The objective of entity-level unlearning is to update $\mathcal{M}$ such that it no longer recalls, generates, or depends on any information from $\mathcal{K}_f$, while maintaining its overall performance on the retained set $\mathcal{P}_r = D \setminus \mathcal{K}_f$.

\smallskip \noindent \textit{Unlearning from the Editing Perspective.} 
In this paper, we frame knowledge unlearning as a specific instance of the model editing task. However, rather than modifying the model's output to produce a new, desired answer, our objective is to suppress specific knowledge by encouraging the model to return uncertain or null responses.
Building on this perspective, the goal of unlearning is to find updated parameters $\theta'$ such that, for any $d \in \mathcal{K}_f$, the model output $\mathcal{M}(d; \theta')$ yields \textit{``Unknown''} (or similar uncertainty-indicating statements). %At the same time,  
Meanwhile, for any $r \in \mathcal{P}_r$, the model should preserve its original behavior, with $\mathcal{M}(r; \theta')$ remaining consistent with its pre-unlearning output. %: $\mathcal{M}(r; \theta')$ should remain consistent with its pre-unlearning output. This is formulated as:
Formally, this objective is expressed as:
\begin{equation*}
    \theta'= \arg\max_{\theta'} ( 
\sum_{d \in \mathcal{K}_f} \mathcal{E}_f(\mathcal{M}(d,\theta')) 
+ 
\sum_{r \in \mathcal{P}_r} \mathcal{E}_r(\mathcal{M}(r,\theta')) 
),
% \label{goal}
\end{equation*}
where $\mathcal{E}_f$ measures the unlearning effectiveness, i.e., \ how well the target knowledge is suppressed, and $\mathcal{E}_r$ evaluates the model's retained capabilities.

% \todo{Give an example to show what is entity unlearning?}

%For example, let the target entity be ``Jackie Chan''. The associated fact set $\mathcal{K}(\text{Jackie Chan})$ might include statements such as:
%$ e_1 $: Jackie Chan is a martial artist and actor,
%$ e_2 $: Jackie Chan was born in Hong Kong,
%$ e_i $: Jackie Chan starred in the movie ``Rush Hour''.

For example, consider the target entity ``Jackie Chan". The associated fact set $\mathcal{K}(\text{Jackie Chan})$ may include facts such as:
$e_1$: Jackie Chan is a martial artist and actor;
$e_2$: Jackie Chan was born in Hong Kong;
$e_i$: Jackie Chan starred in the movie ``Rush Hour".

Entity-level unlearning seeks to suppress this entire set of facts so that the model no longer reproduces any information about ``Jackie Chan'' , regardless of phrasing, while maintaining fluency and correctness on all unrelated content.

\subsection{Autoregressive Language Models}
An \emph{autoregressive language model} $\mathcal{M}$ generates text by predicting the next token $x_t$ conditioned on the preceding tokens $x_{1:t-1} = x_1, \ldots, x_{t-1}$. The model is typically parameterized as an $L$-layer transformer. At each layer $\ell$, the hidden state of the $t$-th token, denoted $h^{(\ell)}_t$, is updated according to:
\begin{equation}
h^{(\ell)}_t = h^{(\ell-1)}_t + a^{(\ell)}_t + m^{(\ell)}_t,
\end{equation}
where:
$a^{(\ell)}_t=\text{Attn}^{(\ell)}\bigl(h^{(\ell-1)}_{1:t}\bigr)$
% \nb{V: What is this notation $h^{(l-1)}_{1:t}$? What is the meaning of $_{1:t}$} 
is the output of the self-attention based on the previous tokens $h_{1:t}$, and
$m^{(\ell)}_t = W^{(\ell)}_{\text{out}}  \sigma \bigl( W^{(\ell)}_{\text{in}} \, \gamma(h^{(\ell-1)}_t) \bigr)$ is the output of the multi-layer perceptron (MLP), with weight matrices $W^{(\ell)}_{\text{in}} \text{ and }  W^{(\ell)}_{\text{out}}$, and  normalization layers $\gamma$ and $\sigma$.

\subsection{Knowledge Location in Language Models} 
\label{loc_section}
Several studies \cite{rome,memit} view each MLP in a transformer as a key–value store encoding  factual associations.
Specifically, the MLP output at layer $\ell$ for token $t$ can be expressed as:
\begin{equation}
   m^{(\ell)}_t = \underbrace{W^{(\ell)}_{\text{out}}}_{\textit{V}} \underbrace{ \sigma \bigl( W^{(\ell)}_{\text{in}} \, \gamma(h^{(\ell-1)}_t) \bigr)}_{\textit{K}}, 
   \label{loc}
\end{equation}
where the $\textit{K}$ encodes the subject–relation context, and $\textit{V}$ retrieves the associated object.  This formulation underpins editing-based methods that target factual memory directly in the MLP weights.

\begin{figure}[t]
    \centering
    \includegraphics[scale=0.25]{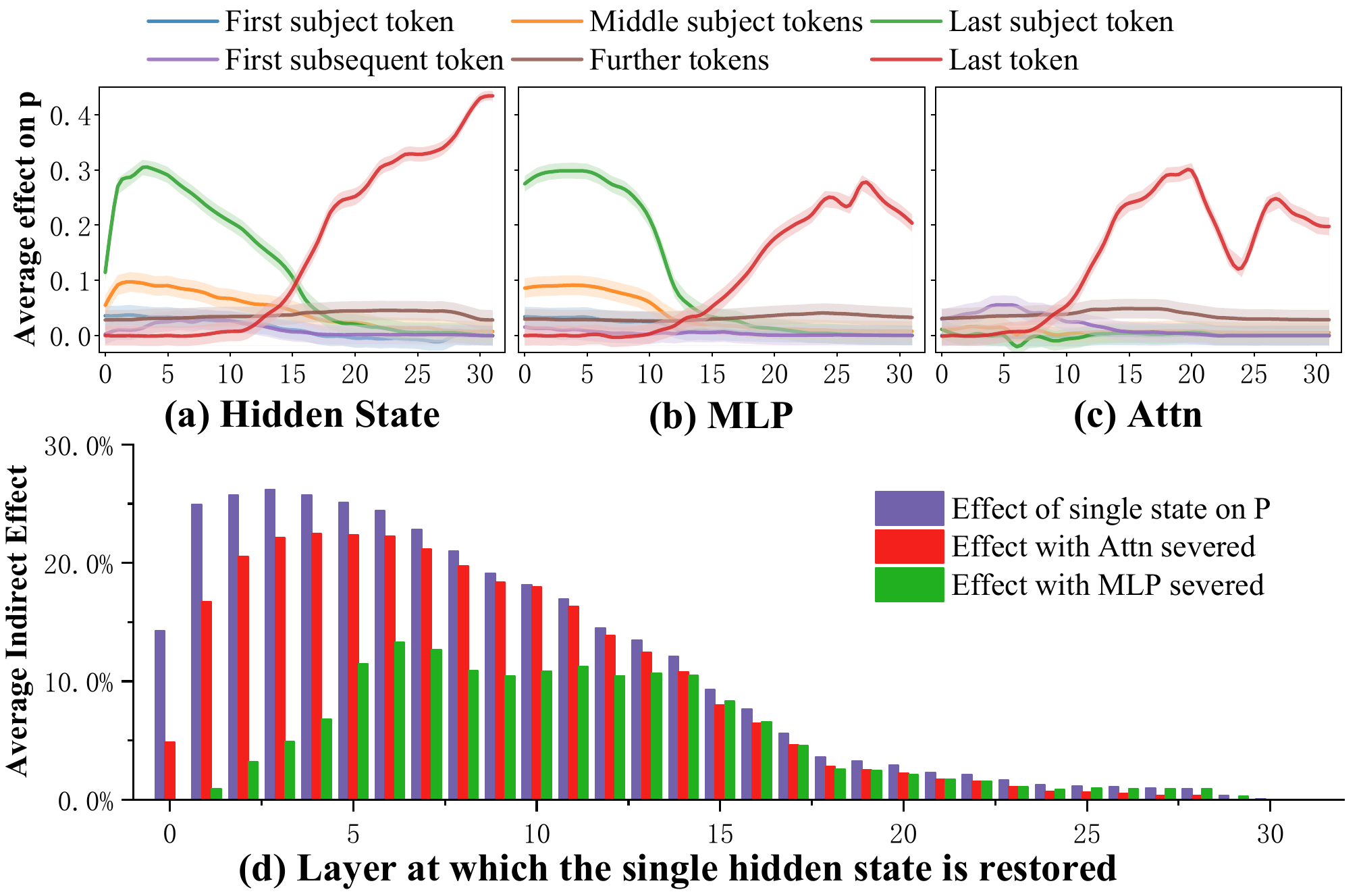}
    \caption{Causal effects to the probability (P)  of model'output.
    (a) Strong causality at a `late site' in the last layers at the last token and strongly causal states at an `early site' in middle layers at the last subject token. 
        (b) MLP contributions dominate the early site. 
        (c) Attention is important at the late site.
        (d) Average indirect causal effect of hidden states on output probability. Disabling MLP components (green) results in a greater reduction in influence compared to attention components (red), indicating that MLP pathways are the primary carriers of entity-level knowledge.}
    \label{locations}
\end{figure}

To investigate where entity-level knowledge resides %is stored 
in LLMs, we perform causal tracing and intervention analyses across 100 entities from RWKU \cite{jinrwku}.
As illustrated in Figure~\ref{locations}(a–c), MLP modules consistently exert a stronger influence on output probabilities than attention modules—especially in higher layers and at the final (prediction) token.
To further support this finding, Figure~\ref{locations}(d) shows a severing experiment, in which disabling MLP modules substantially reduces the model's causal effect, whereas severing attention modules yields a considerably smaller impact.
%Taken together, 
 Collectively, these results suggest that factual knowledge about entities is primarily stored in MLPs, motivating MLP-targeted strategies for entity-level editing and unlearning.

Building on this perspective, we perform low-rank updates to these weight matrices, using rank-one modifications to $W^{(\ell)}_{\mathrm{out}}$ to insert, correct, or delete specific facts without retraining the entire model.  
Unlike prior methods that modify a single or randomly chosen fact, our work aims to remove all information about an entity by identifying and removing the %entire facts associated with same subject.
 the complete set of facts associated with that entity.

\section{Method}

In this section, we introduce a consistency-aware editing method for entity-level unlearning. 

\subsection{Unlearning with Parameters Shift in MLP }

As introduced at Equation~\eqref{loc}, in each individual layer $\ell$, we want to erase all knowledge related to a given entity.
Thus, we denote the input for the $\ell$-th MLP as $K_*$ and the output as $M_*$. Specifically, we define
    $K_r=\{k_1^r, \ldots , k_n^r\} \in  \mathbb{R}^{d\times n}$ as the key matrix used to preserve the model's general behavior, sampled independently from the forget set;   
    %\textcolor{blue}{(not equal to data in  $\mathcal{P}_r$)}\nb{V: What is this? Is it a leftover?}
    $M_r=\{m_1^r, \ldots ,m_n^r\} \in  \mathbb{R}^{d\times n}$ as the associated  value for $K_r$;
    $K_f=\{k_1^f, \ldots, k_u^f\} \in  \mathbb{R}^{d\times u}$ as the key matrix for the forget set related to same entity; and 
    $M_f=\{m_1^f, \ldots ,m_u^f\} \in  \mathbb{R}^{d\times u}$ as the associated  values for $K_f$.
The unlearning objective can then be reformulated as a modification of the $\ell$-th $W_{out}$ parameters $W$ in MLP, by introducing a perturbation $\Delta$ at each layer. The goal is to find $\Delta$ such that:
\begin{equation}
        \Delta = \arg\min_{\widehat{\Delta }} 
        (\|(W+\widehat{\Delta })K_f-M_f\|^2 + \|(W+\widehat{\Delta })K_r-M_r\|^2),
\label{goal}
\end{equation}
where $\|\cdot\|^2$ denotes the  sum of the squared elements in the matrix. The first term enforces the removal of factual associations for the entity, while the second term %preserves the model's behavior.
 ensures the model’s general behavior is preserved.

We can solve Equation~\eqref{goal} by applying the normal equation. To this aim,  we define $R=M_f-WK_f$ and  $WK_r=M_r$.  Then, Equation~\eqref{goal}  can be rewritten as:
\begin{equation}
        \Delta =RK_f^T(K_rK_r^T+K_fK_f^T)^{-1}.
\label{goal2}
\end{equation}
Therefore, to compute $\Delta$, we need to obtain  the residual error $R$ and the key $K_f$ of the unlearning data, along with an estimated key matrix $K_r$ for the retain set. 
In practice, following prior work \cite{rome,memit}, we approximate $K_r K_r^T$ by computing the empirical covariance over a sample of 100,000 random triplets from Wikipedia. 
% Appendix~\ref{prove} provides further details on this procedure.
In the following section, we describe %the construction of the forget keys $K_f$ and the residuals $R$ used for entity unlearning.
how the forget keys $K_f$ and the residuals $R$ are constructed for entity-level unlearning.
% \todo{Does we need to prove it? }

\begin{figure*}[t]
    \centering
    \includegraphics[width=1\textwidth]{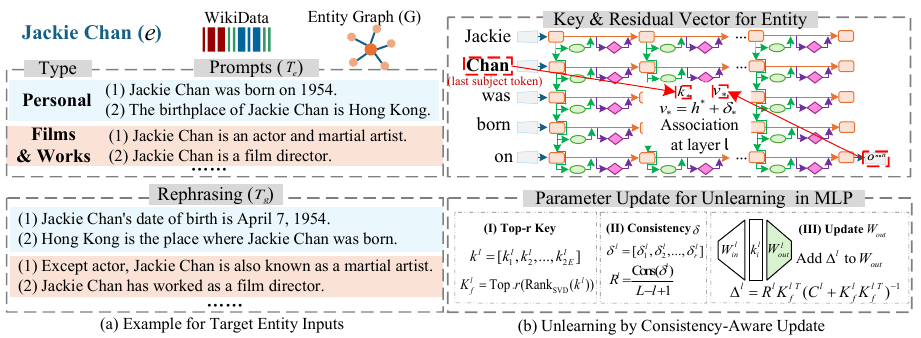}
    \caption{{Overview of our method
     (a) Given an entity $e$ (e.g., Jackie Chan), we retrieve facts from Wikidata and convert them into natural language $T_e$, followed by rephrasing into $T_g$. The union of $T_e$ and $T_g$ is the input of CAE.
    (b) For each prompt, we extract key vectors and optimize a residual $\delta$ at layer $\ell$. 
    The key vectors are selected via SVD-based ranking, while the residuals are regularized using a consistency loss to ensure aligned updates across prompts. We then distribute the residuals $R^\ell$ across subsequent layers.}
    }
    \label{Overview}
\end{figure*}
\subsection{Key Matrix of Entity Knowledge}
%\nb{V: I changed the subsection title. Does it make sense? Maybe no `Representation'? Cause the Key means the Kev-Value's Key. If we add  Representation `Key' maybe the important feature in Representation?}
\label{selection}
Unlike in typical model editing settings where precise input-output pairs are provided for a fact to be modified, %in the entity-level  unlearning setting only the name of the entity along with a few probing questions are given.
 entity-level unlearning is given only the entity name along with a few probing questions. 
 %Therefore, a critical challenge in constructing the key matrix for the forget set associated with an entity is how to generate a diverse and representative set of inputs that sufficiently covers the knowledge associated with that entity.
 Consequently, a key challenge in constructing the key matrix for the forget set is generating a diverse and representative set of inputs that sufficiently capture the entity’s associated knowledge.
To address this, as shown in Figure \ref{Overview}, we first retrieve a range of one-hop facts about the target entity $e$ from Wikidata\footnote{\url{https://query.wikidata.org}}, categorized by their semantic types.
We then convert these structured triples into natural language prompts using predefined templates. This yields a base set $T_e=\{t_1, \ldots ,t_E\}$, where $E$ is the number of inputs.
To further enhance linguistic diversity and improve coverage of the entity's knowledge representation, we apply a text generation model to paraphrase these prompts, resulting in an augmented set $T_g=\{t_1, \ldots, t_E\}$.
% Appendix \ref{dataGeneration} provides details on this procedure.

Our final input set for unlearning is defined as $T_{\mathrm{in}} = T_e \cup T_g$.
For each prompt $x_j \in T_{\mathrm{in}}$ at layer~$\ell$, we compute its corresponding key vector for the $t$-th token as:
\begin{equation}
     k_j^\ell \;=\;\mathrm{key}_\ell(x_j)
     \;=\;\sigma\bigl(W_{\mathrm{in}}^\ell\,\gamma(h_{\ell-1}^j[t])\bigr),
     \label{keys}
\end{equation}
where $\gamma$ represents an intermediate nonlinearity, and $\sigma$ is the activation function of the MLP. 
We focus on the last subject token as the $t$-th token and %we further 
discuss the results on editing last token in Section \ref{token_res}.

However, some prompts in $T_{in}$ may be redundant or induce conflicting updates. 
To select a % compact, informative subset, 
compact and informative subset,
we  compute its SVD, $\mathbf{K}_f^\ell = U \Sigma V^\top$. 
We then project each $k_j^\ell$ onto the top-$r$ left singular vectors $U_{:,1:r}$ and score it by 
$\mathrm{score}(j) = \bigl\|U_{:,1:r}^\top k_j^\ell\bigr\|_2.$
We sort by score and retain the top-$r$ keys whose cumulative projection energy exceeds a threshold $\tau$ (e.g., 95\%). 
The final set $K_f^\ell=\{k_1^\ell, \dots, k_r^\ell\} $ consists of these top-scoring key vectors.

\subsection{Consistency Constrains for Residuals}

In the absence of access to the output value matrix $M_f$, we do not explicitly construct it for %the purpose of entity unlearning.
entity-level unlearning.
Instead, for each prompt $x_j \in T_{\mathrm{in}}$, we optimize a small residual vector $\delta_j \in \mathbb{R}^d$, which is added to the hidden state at layer $L$, denoted by $h_L^j$.
This residual $\delta_j$ is designed to suppress the entity's factual associations without requiring reconstruction of $M_f$.
While the relation $R = M_f - WK_f$ allows us to recover $K_f$, the corresponding output values $M_f$ remain inaccessible.
Consequently, %we modify the hidden representation by applying the residual directly:
we directly modify the hidden representation by applying the residual: $h_L^j + \delta_j$.
A natural starting point is to minimize  the  loss $\mathcal{L}_{\mathrm{NLL}}$ \cite{memit}

\begin{equation}
    \mathcal{L}_{\mathrm{NLL}}
= \frac{1}{r} \sum_{j=1}^r -\log \Pr\bigl(o^{\mathrm{null}}\mid x_j; h_L^j + \delta_j\bigr),
\end{equation}
{where $r$ is the size of $T_{in}$. However, optimizing $\mathcal{L}_{\mathrm{NLL}}$ alone often fails to fully suppress the target fact as some prompts or paraphrases may still elicit the original information.
In practice, the updates induced by individual prompts can be misaligned, pointing in conflicting directions that do not span the full ``knowledge subspace'' associated with the entity.
As a result, certain factual attributes may persist despite the intervention.}

To address this, we add a global consistency constraint that aligns all residuals toward a common update:
\begin{equation}
\mathcal{L}_{\mathrm{Cons}}
= \lambda_{\mathrm{cons}}\,
\Bigl\lVert (h_L^j + \delta_j)
 - \frac{1}{j-1}\sum_{k=1}^{j-1}(h_L^k + \delta_k)\Bigr\rVert_2^2.
 \label{eq7}
\end{equation}
{This penalty encourages consistency across residuals $\delta_j$ by keeping each modified activation $h_L^j + \delta_j$ close to their mean.
In effect, it reduces the variance of updates across the prompt set, preventing individual residuals from conflicting with or negating one another.
As a result, all $\delta_j$ contribute coherently, enabling a more consistent and comprehensive %thorough 
erasure of the target entity's knowledge.}

{Finally, for each $\delta_j$, we  optimize the combined objective:}
\begin{equation}
\delta_j = \arg\min_{\delta}\bigl(\mathcal{L}_{\mathrm{NLL}} + \mathcal{L}_{\mathrm{Cons}}\bigr).
\end{equation}
{By coordinating updates in this manner, we achieve a more comprehensive and uniform erasure of entity knowledge across all prompts. Following optimization, we collect the edited representations and their corresponding keys at layer $L$ into the matrix $R = \bigl\{\delta_1,;\dots,; \delta_r\bigr\}$, which serves as the value component in the weight update. The resulting parameter shift for layer $L$ is then computed using Equation~\eqref{goal2}.}

\subsection{Multi‐Layer Key Extraction and Weight Updates}
{The procedure above outlines how to update a single MLP layer. However, modifying one layer inevitably influences all subsequent activations. To propagate the editing effect and ensure consistent suppression of the entity's influence in higher layers,
we follow \cite{memit} and apply updates with:}
% in a bottom-up, layer-wise manner:}

\smallskip \noindent\textbf{Residual Distribution.}  
{We distribute residual  $\delta_j$ over the remaining layers $\{\ell, \ell+1, \ldots, L\}$: % 
\begin{equation}
     r_j^\ell = \frac{\delta_j}{L - \ell + 1}, 
     \qquad
     R_f^\ell = \{\,r_1^\ell, \dots, r_r^\ell\,\}.
\end{equation}}
\noindent \textbf{Closed‑Form Update.}  
{Finally, at layer $\ell$, we compute the parameter shift using  Equation~\eqref{goal2} %.   We then 
and subsequently update the layer's weights:}  
\begin{equation}
     W_{\mathrm{out}}^\ell \;\leftarrow\; W_{\mathrm{out}}^\ell + \Delta^\ell.
\end{equation}
{By iterating this process from $\ell = 1$ to $L$, we progressively erase the target entity's memory across all layers, while %minimally disrupting the model's remaining behavior. 
preserving the model’s behavior on unrelated content.}

\begin{table*}[ht]
% \tiny
\resizebox{1.99\columnwidth}{!}{
\begin{tabular}{ccccc|ccc|cc|ccccc|cc}
\toprule
 & \multicolumn{4}{c}{Forget Set $\downarrow$} & \multicolumn{3}{c}{Neighbor Set $\uparrow$} & \multicolumn{2}{c}{MIA Set} & \multicolumn{5}{c}{Utility Set$\uparrow$}   &    \multicolumn{2}{c}{Average $\uparrow$ }   \\
\midrule
              Method          & FB     & QA    & AA    & All   & FB        & QA        & All      & FM $\uparrow$            & RM $\downarrow$          & Gen  & Rea  & Tru  & Fac  & Flu   & Mean & Mean\_FN  \\ \midrule
\multicolumn{16}{c}{Llama3-Instruct (8B)}                                                                                                                                        \\\toprule
Before                  & 85.9   & 76.4  & 77.7  & 79.6  & 95.6      & 85.3      & 90.7     & 226.7        & 230.4        & 65.7 & 42.3 & 36.8 & 53.5 & 705.8 & 111.5 & 55.6     \\
ICU                     & 26.2   & 1.9   & 10.3  & 12.8  & 65.0      & 46.5      & 55.7     & 247.1        & 258.4        & 63.6 & 39.3 & 36.4 & 48.2 & 705.0 & 128.5 & 71.5     \\
RepE                    & 29.8   & 33.6  & 37.8  & 34.8  & 46.2      & 38.8      & 42.6     & 292.0        & 290.0        & 64.8 & 26.3 & 37.6 & 17.9 & 703.7 & 105.5 & 53.9     \\ 
\hline
% GA* (Full)              & 40.7   & 36.5  & 43.7  & 41.4  & 68.6      & 68.6      & 68.1     & 1640.9       & 766.2        & 65.5 & 39.7 & 37.8 & 41.9 & 692.4 & 137.4 & 63.4     \\
% GA* (LoRA)              & 70.3   & 65.6  & 67.8  & 68.2  & 80.6      & 75.5      & 77.5     & 879.5        & 665.1        & 64.0 & 37.8 & 37.3 & 43.8 & 711.3 & 119.9 & 54.7     \\
GA (Full)               & 39.1   & 31.6  & 46.7  & 41.9  & 84.6      & 73.6      & 79.0     & 258.6        & 231.0        & 64.9 & 42.0 & 35.9 & 52.5 & 705.1 & 123.7 & 68.6     \\
GA (LoRA)               & 67.0   & 53.2  & 61.8  & 61.3  & 90.1      & 80.4      & 85.3     & 224.1        & 221.6        & 64.7 & 41.5 & 36.6 & 52.8 & 697.3 & 116.6 & 62.0     \\
DPO (Full)              & 46.3   & 38.5  & 41.6  & 41.9  & 59.2      & 51.3      & 55.2     & 243.6        & 240.8        & 64.1 & 42.0 & 31.5 & 25.8 & 725.9 & 109.9 & 56.7     \\
DPO (LoRA)              & 75.3   & 65.4  & 68.6  & 69.5  & 90.0      & 81.5      & 85.6     & 228.0        & 231.2        & 65.6 & 42.0 & 34.5 & 55.5 & 702.7 & 114.3 & 58.1     \\
NPO (Full)              & 33.4   & 21.0  & 24.8  & 26.2  & 76.0      & 69.9      & 72.6     & 278.9        & 263.2        & 64.8 & 41.5 & 34.9 & 41.2 & 712.2 & 125.5 & 73.2     \\
NPO (LoRA)              & 75.1   & 64.3  & 69.0  & 69.7  & 91.3      & 82.2      & 86.7     & 225.1        & 227.0        & 64.9 & 41.7 & 36.0 & 54.0 & 707.3 & 114.2 & 58.5     \\
RT (Full)               & 72.7   & 13.4  & 22.8  & 33.1  & 86.9      & 45.6      & 67.4     & 222.7        & 226.6        & 65.4 & 41.4 & 34.9 & 59.3 & 588.1 & 122.7 & 67.2     \\
RT (LoRA)               & 85.4   & 49.6  & 53.2  & 60.5  & 87.3      & 74.1      & 81.9     & 226.0        & 223.9        & 64.5 & 41.2 & 33.6 & 58.2 & 667.0 & 115.9 & 60.7     \\ \hline
MEMIT                   & 29.7   & 18.6  & 31.4  & 26.6  & 80.7      & 72.0      & 76.4     & 243.0        & 230.0        & 66.0 & 40.0 & 40.0 & 52.8 & 709.0 & 130.3 & 74.9     \\
AlphaEdit               & 62.7   & 52.6  & 58.9  & 58.1  & 85.1      & 83.0      & 84.1     & 233.0        & 230.0        & 65.6 & 40.4 & 39.6 & 53.0 & 708.0 & 119.0 & 63.0     \\
EMMET                   & 28.2   & 22.6  & 36.1  & 29.0  & 82.7      & 77.0      & 79.9     & 242.0        & 230.0        & 65.7 & 39.6 & 40.2 & 52.3 & 708.0 & 129.7 & 75.4     \\
\rowcolor{gray!20} CAE                    & 18.2   & 7.4   & 22.0  & 15.9  & 79.8      & 64.5      & 72.2     & 275.0        & 230.0        & 65.4 & 38.7 & 40.4 & 52.1 & 708.0 & 133.4 & \textbf{78.1}     \\
\toprule
\multicolumn{16}{c}{Llama3.1-Instruct (8B)}                                                                                                                                         &          \\
\toprule
Before                  & 67.5   & 68.1  & 68.1  & 67.9  & 82.0      & 77.3      & 79.7     & 215.0        & 219.0        & 66.1 & 45.2 & 39.5 & 55.3 & 694.0 & 115.8 & 55.9     \\
ICU                  & 21.8   & 5.3   & 9.0   & 12.0  & 32.6      & 6.1       & 19.4     & 237.0        & 254.0        & 64.1 & 27.0 & 37.7 & 36.8 & 695.0 & 114.5 & 53.7     \\ \hline
GA (Full)                & 37.2   & 29.7  & 43.6  & 36.9  & 77.2      & 74.1      & 75.6     & 248.3        & 219.8        & 65.9 & 42.4 & 39.7 & 55.4 & 689.2 & 126.8 & 69.4     \\
GA (LoRA)                & 60.2   & 55.8  & 61.4  & 59.1  & 79.8      & 76.4      & 78.1     & 223.7        & 221.5        & 65.5 & 40.9 & 39.2 & 55.8 & 684.0 & 117.4 & 59.5     \\
DPO (Full)               & 39.8   & 32.3  & 36.4  & 36.2  & 54.7      & 47.6      & 51.1     & 234.9        & 229.1        & 65.6 & 42.9 & 32.4 & 20.1 & 727.1 & 110.5 & 57.5     \\
NPO (Full)               & 27.6   & 18.6  & 21.5  & 22.5  & 67.4      & 64.6      & 66.0     & 249.8        & 226.4        & 65.6 & 41.3 & 38.5 & 53.5 & 674.6 & 128.9 & 71.7     \\
NPO (LoRA)               & 64.3   & 62.4  & 64.6  & 63.8  & 79.8      & 76.3      & 78.0     & 217.5        & 219.0        & 65.7 & 41.6 & 39.5 & 54.8 & 688.6 & 115.6 & 57.1     \\
% passage\_full           & 31.8   & 28.1  & 33.9  & 31.3  & 62.2      & 63.2      & 62.7     & 2031.9       & 1005.9       & 65.1 & 35.9 & 40.0 & 36.2 & 685.4 & 144.1 & 65.7     \\
% passage\_lora           & 66.1   & 66.7  & 67.0  & 66.6  & 80.8      & 76.6      & 78.7     & 243.3        & 232.9        & 66.1 & 42.8 & 39.5 & 52.5 & 698.6 & 115.1 & 56.0     \\
RT (Full)                & 16.8   & 15.0  & 12.7  & 14.8  & 24.3      & 38.4      & 31.3     & 215.3        & 218.3        & 65.7 & 40.6 & 38.8 & 29.9 & 569.6 & 114.2 & 58.2     \\
RT (LoRA)                & 54.5   & 50.9  & 47.5  & 50.9  & 65.3      & 67.7      & 66.5     & 215.2        & 218.3        & 65.6 & 41.6 & 38.5 & 42.8 & 653.1 & 113.4 & 57.8     \\ \hline
MEMIT                  & 30.5 & 23.6 & 41.6 & 31.9 & 74.5 & 69.8 & 72.2 & 228.0 & 219.0 & 65.8 & 47.1 & 38.3 & 56.8 & 693.0 & 129.3 & 70.1 \\
AlphaEdit              & 64.6 & 58.8 & 63.8 & 62.4 & 82.0 & 76.9 & 79.5 & 216.0 & 219.0 & 66.0 & 45.5 & 39.6 & 55.5 & 695.0 & 118.1 & 58.5 \\
EMMET                    & 36.1 & 29.0 & 45.2 & 36.8 & 77.6 & 71.9 & 74.8 & 225.0 & 219.0 & 65.8 & 45.6 & 39.0 & 55.4 & 695.0 & 127.3 & 69.0 \\
\rowcolor{gray!20} CAE                    & 18.7   & 12.4  & 28.4  & 19.8  & 64.5      & 64.8      & 64.7     & 257.0        & 219.0        & 66.0 & 45.1 & 39.6 & 55.2 & 695.0 & 132.4 & \textbf{72.4} \\
\bottomrule
\end{tabular}
}
\caption{Overall performance comparison across unlearning methods on RWKU (100 subjects).
Mean is a weighted average of Forget (0.4), Neighbor (0.2), Utility (0.3), and MIA (0.1).
All is the average value of FB, QA and AA.
Mean$\_$FN is the simple average of Forget (All) and Neighbor (All). 
% Appendix \ref{Experimentss} gives more details of each metrics.
}
\label{mains}
\end{table*}

\section{Experiments}
\label{Experiments}

{To evaluate the effectiveness of CAE,  %we aim to explore the fol lowing research questions:
we investigate the following research questions:
\begin{itemize}[itemsep=0.5ex, leftmargin=5mm]
% \item
% \textbf{RQ1:} How does CAE perform on RWKU and ToFU benchmarks compared to existing baselines?  
% \item
% \textbf{RQ2:} How robust is CAE across different types of forgetting questions and entity categories? 
% \item
% \textbf{RQ3:} How do the SVD-based key selection and consistency loss affect CAE's effectiveness?  
\item \textbf{RQ1:} How does CAE perform on standard unlearning benchmarks compared with existing editing-, training-, and prompt-based baselines?

\item \textbf{RQ2:} How robust is CAE across diverse unlearning scenarios, including different entities, question types, and sequential multi-entity forgetting?

\item \textbf{RQ3:} %How do data scale, edit count, and token-level choices influence the effectiveness and side effects of CAE?
How do factors such as data scale, number of edits, and token-level selection affect the effectiveness and side effects of CAE?

\item \textbf{RQ4:} How do the internal mechanisms of CAEs, particularly the consistency constraint, underpin edits that are stable, leakage-resistant, and geometrically coherent?

\item \textbf{RQ5:} How well does CAE generalize across model architectures and alternative data sources, such as LLM-generated synthetic examples?
\end{itemize}}

\subsection{Experimental Setup}
\noindent \textit{Datasets and Models.}
We evaluate forgetting performance on two entity-level benchmarks: RWKU \cite{jinrwku} and ToFU \cite{mainitofu}, %which respectively contain 100 and 20 entities targeted for removal. 
which contain 100 and 20 entities targeted for removal, respectively.
We focus on the single-entity unlearning setting, where each experiment targets the removal of one entity at a time. 
Final results are reported as the average performance across all individual unlearning cases.

%It is worth noting that we mainly focus on the single-entity unlearning setting, where each experiment targets the removal of one entity at a time. 
%And we also test the sequential unlearning performance to show the effectiveness. 

It is worth noting that our primary focus is on the single-entity unlearning setting, where each experiment removes one entity at a time. We also evaluate sequential unlearning to demonstrate the method’s effectiveness in handling multiple, successive entity removals.

This %results in 
yields
100 and 20 separately edited models for RWKU and ToFU, respectively, and we evaluate the performance of each individually. 
For ToFU, we evaluate on the 10\% forget set.
Final results are reported as the average across all unlearning instances.

Experiments are conducted on LLaMA3-Instruct (8B) and LLaMA3.1-Instruct (8B) from HuggingFace. LLaMA2-7B-Chat is a fine-tuned model on the ToFU dataset.\footnote{\url{https://huggingface.co/open-unlearning/tofu_Llama-2-7b-chat-hf_full}}
All editing methods are executed %run 
on two NVIDIA A100 GPUs (40GB each). 
Since RWKU does not provide official results for LLaMA3.1-Instruct (8B), we additionally train this model using four NVIDIA A800 GPUs (80G).
%The $\lambda_{cons}$ we used in Eq. \ref{eq7} is 0.05.
The consistency weight $\lambda_{cons}$
 in Eq.~\ref{eq7} is set to 0.05.

\smallskip \noindent \textit{Evaluation Metrics.} 
{For RWKU, following previous work \cite{jinrwku}, we evaluate unlearning performance from four perspectives. }
{(1) %We assess the 
\emph{Forget set}: %, which 
includes \emph{fill-in-the-blank (FB)} probes, \emph{question-answer (QA)} probes, and \emph{adversarial attack (AA)} probes, all related to the target entity. 
(2) %We examine the 
\emph{Neighbor set}: %, which 
contains facts closely related to but not entirely belonging to the target entity, measured through FB and QA probes.
(3) %We employ 
\emph{Membership inference attacks (MIA)}: Compares %by comparing 
the model's predictions on the \emph{forget member (FM)} set versus %and 
the \emph{Retain Member (RM)} set, %to rigorously audit 
rigorously auditing whether the model still retains the target knowledge.
(4) \emph{Model utility}: Assesses the model’s overall performance after unlearning using general-purpose benchmarks, %We assess the model's utility after unlearning using general-purpose benchmarks, 
including MMLU (Gen) \cite{hendrycks2021measuring}, BBH (Rea) \cite{suzgun2023challenging}, TruthfulQA (Tru) \cite{lin2022truthfulqa}, TriviaQA (Fac) \cite{joshi-etal-2017-triviaqa}, and AlpacaEval (Flu)\cite{alpaca_eval}.}
{For ToFU, following \cite{ma2025unveiling}, \emph{forgetting} is measured using prediction probability and ROUGE scores. 
Post-unlearning capabilities are evaluated using \emph{model utility}, which includes the \emph{retain set score (RS)}, \emph{real authors set score (RAS)}, and \emph{world facts score (WFS).} }

\smallskip \noindent \textit{Baselines } 
{We compare CAE with a comprehensive set of unlearning methods. 
These include \emph{in-context unlearning (ICU)} \cite{pawelczyk2024context}, which achieves unlearning without changing model weights, and \emph{representation engineering (RepE)} \cite{li2024wmdp}, which perturbs hidden activations using control vectors. 
We also include \emph{gradient ascent (GA)} \cite{jang-etal-2023-knowledge}, which explicitly maximizes loss on the forget set.
For preference-based approaches, we evaluate \emph{direct preference optimization (DPO)} \cite{rafailov2023direct} and \emph{negative preference optimization (NPO)} \cite{zhang2024negative}, as well as \emph{rejection tuning (RT)}, which fine-tunes the model to reject responses related to the target entity.} 
{We additionally include enhanced variants such as \emph{gradient difference (Grad. Diff.)} \cite{liu2022continual} and \emph{KL minimization (KL Min.)}, which aim to improve unlearning efficacy through more precise loss shaping.
Finally, for editing-based approaches, we focus on locate-then-edit methods and compare with MEMIT~\cite{memit}, EMMET~\cite{gupta2024unified}, and AlphaEdit~\cite{fang2024alphaedit}. 
% Details of experimental setups are provided in Appendix~\ref{Experimentss}.}
This set covers a broad spectrum of strategies, from activation perturbation and loss-based methods to preference and editing-based techniques, allowing for a thorough comparison with CAE.

\subsection{Main Results}
To address \textbf{RQ1}, we conduct experiments on the RWKU and ToFU. 
%And w
We find CAE achieves state-of-the-art performance on both benchmarks, %RWKU and ToFU
outperforming all editing-based baselines and many training-based methods. 
%It delivers stronger forgetting while preserving neighbor knowledge and model utility more effectively than prior approaches, demonstrating a superior forgetting–preservation trade-off.
It achieves stronger forgetting while better preserving neighbor knowledge and overall model utility, demonstrating a superior trade-off between forgetting and retention compared with prior approaches.

\subsubsection{Results on RWKU}

For RWKU, following the experimental setup of %the setting by 
\cite{jinrwku}, we evaluate each method's performance on %with 
100 subjects.

\textbf{For forgetting performance, CAE delivers the most effective and comprehensive removal of entity knowledge. }
As shown in Table \ref{mains}, CAE achieves an All score of 15.9, outperforming %prompt-based methods, training-based methods, as well as editing-based approaches.
prompt-based, training-based, and other editing-based approaches.
%CAE not only achieves a comparable forgetting score (78.1 and 72.4 on Mean\_FN) but does so through direct model editing, \textit{without requiring prompt-level interventions}. 
Importantly, CAE attains competitive forgetting scores (78.1 and 72.4 on Mean\_FN) through direct model editing, \textit{without relying on prompt-level interventions}.
Compared to training-based approaches such as GA and DPO, CAE makes a significant improvement on Mean\_FN by at least 2–5\%, exhibiting  stronger entity erasure with substantially lower computational overhead (see Section \ref{costs}) and without retraining on large-scale data.
Moreover, CAE outperforms other editing methods on Mean\_FN, demonstrating its ability to overcome the inconsistency limitations of these approaches.
Notably, while ICU achieves strong forgetting, it relies heavily on prompt manipulation, often at the expense of generality and controllability, which leads to weaker performance on neighboring tasks. %hence its lower performance on neighboring tasks. 
In contrast, CAE consistently outperforms most baselines on metrics such as FB, QA, and AA,  highlighting its robust and reliable entity-unlearning capability.

We further evaluate privacy via MIA on FM and RM.
CAE achieves both the highest FM and lowest RM, thereby producing the largest FM–RM gap among all methods. 
This pronounced gap demonstrates CAE's ability to selectively erase target knowledge without degrading the model's overall factual understanding.

%In summary, CAE achieves the best forgetting performance among editing-based methods, while also surpassing prompt-based and training-based baselines in both effectiveness and efficiency for entity unlearning.

In summary, CAE not only delivers the strongest forgetting performance among editing-based methods but also surpasses prompt-based and training-based baselines in both effectiveness and efficiency for entity unlearning.

\textbf{Beyond effective forgetting and privacy protection, CAE also excels in preserving related knowledge and overall model utility. }
{In the neighbor set, CAE performs better than prompt-based methods like ICU and RepE, improving by 20\%-40\%.  }
{For utility, CAE retains performance across all metrics compared to the pre-unlearning model and most other methods. 
In contrast, %However, 
ICU, RepE, and DPO (Full) exhibit a significant decline in Fac scores across both models, revealing their imbalance between unlearning effectiveness and utility retention.}

These results indicate that CAE not only removes target knowledge with high precision but also minimizes unintended degradation of related factual content, while preserving the fluency, readability, and informativeness of the model's outputs. 

To provide a comprehensive view of unlearning performance, we report two metrics: Mean and Mean$\_$FN. 
The Mean score reflects a method's ability to perform well across the full spectrum of requirements, rewarding balanced solutions rather than extreme optimization of a single dimension. 
In contrast, \textit{Mean\_FN} focuses specifically on the trade-off between target forgetting and preservation. 
It averages Forget (All) and Neighbor (All) scores, thus directly measuring how well a method can remove target knowledge while minimizing unintended side effects on semantically related content.
CAE  obtains the highest Mean$\_$FN, demonstrating its ability to forget precisely while preserving useful context.
In summary, CAE offers the strongest overall performance across all dimensions of the unlearning objective, achieving an effective balance between forgetting and knowledge retention. 

% \todo{why need 10, casue some method only running on 10 subjects. Just follow other paper's setting.}

Following \cite{yuan2025towards}, we evaluate unlearning performance across 10 subjects.
%As shown in Table \ref{10subs}, our CAE method consistently outperforms all other approaches, achieve 12.9/12.4 forget performance (All for forget), own the best results on Llama3-Instruct, even when adversarial methods (ADV) are applied. 
As shown in Table \ref{10subs}, CAE consistently outperforms all other methods, achieving forget performance scores of 12.9/12.4 (All for Forget) and delivering the best results on Llama3-Instruct, even under adversarial conditions (ADV).
In contrast, while GA and NPO achieve improved forgetting with ADV, their performance on neighbor knowledge preservation noticeably degrades, e.g. after use Adv, the Neighbor performace (All for Neighbor) drops by %has experienced a 
10-20\%. % decline.
%In summary, CAE get the best Mean value 111.0 on Llama3.1-Instruct and the second performance on Llama3-Instruct, and considering the overall cost of forgetting, CAE achieves a better balance between forgetting effectiveness and side effects.
Overall, CAE attains the highest Mean value of 111.0 on Llama3.1-Instruct and the second-highest on Llama3-Instruct. Considering both effectiveness and computational cost, CAE achieves the most balanced trade-off between precise forgetting and minimizing side effects.

\begin{table*}[]
\centering
% \resizebox{1.99\columnwidth}{!}{
\begin{tabular}{ccccc|ccc|cc|cccc|c}
\toprule
          & \multicolumn{4}{c|}{Forget} & \multicolumn{3}{c|}{Neighbor} & \multicolumn{2}{c|}{MIA} & \multicolumn{4}{c|}{Utility} &       \\
\midrule
Method    & FB    & QA   & AA   & All  & FB       & QA      & All     & FM         & RM         & Rea   & Tru  & Fac  & Flu   & Mean  \\
\midrule
\multicolumn{15}{c}{Llama3-Instruct(8B)}           \\
\midrule
Before    & 85.6  & 70.3 & 74.7 & 76.9 & 93.1     & 82.0    & 87.6    & 236.5      & 230.9      & 41.0  & 36.4 & 53.7 & 704.6 & 91.9  \\
ICU    & 28.5  & 4.2  & 12.3 & 15.0 & 69.5     & 50.2    & 59.9    & 243.0      & 259.0      & 37.1  & 37.5 & 48.6 & 702.0 & 109.0 \\
\toprule
GA        & 72.0  & 64.6 & 68.5 & 68.4 & 85.0     & 74.7    & 79.9    & 241.4      & 234.6      & 40.4  & 37.6 & 49.6 & 710.3 & 93.0  \\
GA (Adv)     & 63.0  & 48.2 & 60.5 & 57.2 & 75.8     & 72.1    & 74.0    & 202.0      & 176.5      & 40.1  & 35.2 & 49.4 & 717.0 & 94.6  \\
GA (GDR)     & 72.6  & 64.0 & 69.7 & 68.8 & 86.2     & 76.5    & 81.4    & 242.8      & 236.8      & 39.6  & 36.8 & 50.4 & 710.3 & 92.9  \\
GA (Adv-GDR)  & 69.2  & 52.4 & 66.1 & 62.6 & 85.7     & 73.7    & 79.7    & 205.2      & 184.5      & 41.4  & 35.4 & 50.5 & 712.1 & 94.4  \\
GA (KLR)     & 70.7  & 57.5 & 69.9 & 66.0 & 80.5     & 70.5    & 75.5    & 242.4      & 230.8      & 41.5  & 35.6 & 54.0 & 704.4 & 93.9  \\
GA (Adv-KLR)  & 58.8  & 43.8 & 59.5 & 54.0 & 76.9     & 63.0    & 70.0    & 371.3      & 340.8      & 41.2  & 33.8 & 50.5 & 712.6 & 98.5  \\
NPO       & 46.6  & 39.0 & 35.3 & 40.3 & 79.2     & 70.9    & 75.1    & 263.3      & 241.4      & 40.5  & 36.0 & 56.7 & 695.9 & 104.8 \\
NPO (Adv)   & 19.7  & 14.7 & 12.0 & 15.5 & 67.0     & 59.7    & 63.4    & 270.1      & 238.9      & 39.3  & 34.0 & 56.8 & 663.1 & 110.5 \\
NPO (GDR)    & 52.2  & 43.9 & 42.9 & 46.3 & 82.5     & 70.5    & 76.5    & 254.5      & 240.1      & 39.6  & 37.2 & 51.4 & 708.2 & 101.4 \\
NPO (Adv-GDR) & 25.5  & 22.1 & 16.5 & 21.4 & 71.9     & 69.1    & 70.5    & 248.8      & 223.1      & 41.9  & 35.8 & 52.4 & 705.2 & 110.5 \\
NPO (KLR)    & 52.5  & 40.6 & 43.2 & 45.4 & 83.2     & 72.1    & 77.7    & 253.0      & 236.9      & 40.9  & 35.4 & 54.2 & 704.9 & 102.6 \\
NPO (Adv-KLR) & 23.6  & 18.9 & 16.0 & 19.5 & 72.1     & 66.8    & 69.5    & 347.2      & 318.1      & 41.7  & 35.6 & 55.3 & 697.1 & 113.4 \\
\toprule
MEMIT     & 25.9  & 14.4 & 34.8 & 25.0 & 70.4     & 54.7    & 62.6    & 249.0      & 231.0      & 41.4  & 36.2 & 53.4 & 705.0 & 107.7 \\
AlphaEdit & 63.7  & 48.5 & 62.4 & 58.2 & 84.8     & 76.4    & 80.6    & 231.0      & 238.0      & 42.9  & 36.8 & 55.2 & 705.0 & 99.2  \\
EMMET     & 33.8  & 19.8 & 38.9 & 30.8 & 75.4     & 64.0    & 69.7    & 231.0      & 246.0      & 41.1  & 36.0 & 53.5 & 703.0 & 106.6 \\
\rowcolor{gray!20} CAE      & 10.2  & 9.5  & 19.1 & 12.9 & 55.7     & 51.2    & 53.5    & 318.0      & 232.0      & 40.5  & 36.4 & 53.3 & 704.0 & 111.2 \\
\toprule
\multicolumn{15}{c}{Llama3.1-Instruct(8B)}                                                                                            \\
\midrule
Before    & 63.9  & 65.1 & 69.5 & 66.2 & 74.1     & 69.8    & 72.0    & 223.5      & 218.2      & 42.2  & 35.4 & 61.2 & 695.2 & 94.8  \\
ICU    & 22.0  & 5.0  & 9.0  & 12.0 & 32.2     & 6.2     & 19.2    & 238.0      & 254.0      & 27.1  & 37.6 & 36.9 & 695.0 & 95.3  \\
\toprule
GA        & 50.7  & 45.4 & 61.2 & 52.4 & 45.6     & 37.2    & 41.4    & 248.9      & 241.9      & 43.2  & 35.8 & 48.7 & 726.6 & 92.3  \\
GA (Adv)     & 32.0  & 22.5 & 36.0 & 30.2 & 27.5     & 21.0    & 24.3    & 173.7      & 125.9      & 39.8  & 33.0 & 28.8 & 730.1 & 88.2  \\
GA (GDR)  & 55.4  & 49.6 & 63.9 & 56.3 & 60.2     & 53.5    & 56.9    & 239.8      & 231.3      & 44.2  & 35.0 & 53.9 & 718.5 & 95.0  \\
GA (Adv-GDR)  & 44.0  & 34.1 & 47.8 & 42.0 & 62.6     & 52.5    & 57.6    & 71.9       & 62.3       & 43.2  & 35.8 & 52.7 & 718.6 & 97.1  \\
GA (KLR)     & 62.7  & 49.9 & 66.4 & 59.7 & 67.9     & 61.2    & 64.6    & 235.8      & 223.0      & 42.6  & 35.4 & 59.0 & 682.1 & 95.2  \\
GA (Adv-KLR)  & 50.8  & 42.0 & 54.8 & 49.2 & 59.8     & 59.8    & 59.8    & 69.1       & 67.2       & 43.1  & 33.4 & 57.3 & 697.5 & 94.7  \\
NPO       & 35.7  & 40.2 & 39.0 & 38.3 & 67.3     & 66.2    & 66.8    & 241.4      & 220.5      & 42.5  & 35.6 & 61.8 & 684.2 & 105.1 \\
NPO (Adv)   & 18.0  & 21.7 & 16.5 & 18.7 & 60.0     & 57.2    & 58.6    & 108.3      & 86.9       & 41.1  & 35.4 & 61.4 & 677.8 & 107.9 \\
NPO (GDR)  & 42.4  & 37.2 & 42.0 & 40.5 & 74.0     & 66.7    & 70.4    & 236.3      & 220.1      & 43.0  & 35.4 & 60.8 & 698.8 & 105.1 \\
NPO (Adv-GDR) & 23.1  & 20.8 & 16.7 & 20.2 & 62.4     & 59.7    & 61.1    & 91.0       & 77.6       & 42.6  & 35.4 & 60.7 & 696.1 & 108.3 \\
NPO (KLR) & 40.6  & 41.4 & 42.2 & 41.4 & 73.3     & 69.9    & 71.6    & 234.4      & 218.8      & 42.3  & 35.4 & 61.5 & 695.1 & 104.9 \\
NPO (Adv-KLR) & 24.1  & 18.5 & 19.4 & 20.7 & 65.0     & 61.0    & 63.0    & 88.9       & 74.9       & 42.2  & 35.2 & 60.5 & 690.2 & 108.0 \\
\toprule
MEMIT     & 26.7  & 23.3 & 38.1 & 29.4 & 67.0     & 59.6    & 63.3    & 239.0      & 219.0      & 44.6  & 37.8 & 54.3 & 693.0 & 107.3 \\
AlphaEdit & 61.7  & 58.6 & 65.7 & 62.0 & 74.2     & 75.2    & 74.7    & 225.0      & 218.0      & 44.1  & 38.8 & 54.2 & 692.0 & 96.5  \\
EMMET     & 29.7  & 26.3 & 42.1 & 32.7 & 68.6     & 60.6    & 64.6    & 236.0      & 219.0      & 44.4  & 37.6 & 54.3 & 695.0 & 106.1 \\
\rowcolor{gray!20} CAE      & 11.6  & 8.6  & 17.1 & 12.4 & 51.5     & 44.7    & 48.1    & 273.0      & 219.0      & 43.2  & 38.6 & 53.6 & 694.0 & 111.0 \\
\bottomrule
\end{tabular}
\caption{Results on 10 subject. GA(ADV) and NPO(ADV) are enhanced training methods based on Latent Adversarial Unlearning, while GDR and KLR refer to auxiliary training strategies that leverage Gradient Difference (Grad. Diff.) and KL Minimization (KL Min.), respectively.}
\label{10subs}
\end{table*}

\subsubsection{Results on ToFU}
{For ToFU, we evaluate performance with 20 subjects on LLaMA2‑7B‑Chat.
\textbf{CAE demonstrates the best trade-off between forgetting effectiveness and model utility.}
As shown in Table \ref{tofuss_d}, CAE achieves a balanced ranking across all methods: obtaining 72.41\% on h-mean, second only to %only lower than 
Pref.OPT.}
% although CAE's Prob. score and ROUGE are not the lowest among all methods, it achieves notably higher retained accuracy on the Retain Set while maintaining competitive RAS and WFS scores,  \nb{provide some specific results here}.
{These strong retention metrics lead to a Model Utility that surpasses most baselines such as GA and KL Min., resulting in a superior harmonic mean.}
{This suggests that CAE not only successfully removes targeted knowledge using just 20 examples per entity but also preserves unrelated world knowledge. 
While some methods (e.g., Pref. Opt.) achieve slightly higher RS scores, they do so at the cost of extreme degradation in forgetting metrics, indicating insufficient erasure. 
}
{Compared to other editing methods, CAE outperforms %them 
on most metrics, with only a slight drop in Model Utility compared to EMMET, further demonstrating the advantage of consistency constraints for entity unlearning.}
{Overall, CAE provides a balanced and robust unlearning solution under the ToFU setting.}

\begin{table*}[h]
\centering
\begin{tabular}{ccc|cccc|c}
\toprule
Method      & Prob.  & ROUGE  & RS  & RAS  & WFS  & Model Utility & h-mean  \\
\midrule
Before     & 0.9908 & 0.9793 & 0.8737   & 0.5893    & 0.5308    & 0.6349        & 0.0309  \\
Grad. Ascent & 0.0009 & 0.2319 & 0.6803   & 0.5533    & 0.5020    & 0.5694        & 0.6603  \\
Grad. Diff.  & 0.1237 & 0.3717 & 0.8326   & 0.6584    & 0.5899    & 0.6795        & 0.6995  \\
KL Min.      & 0.0002 & 0.1110 & 0.6203   & 0.5448    & 0.5104    & 0.5549        & 0.6638 \\
Pref. Opt.   & 0.3486 & 0.0147 & 0.9024   & 0.6777    & 0.6349    & 0.7213        & 0.7453 \\
NPO-GD       & 0.0344 & 0.2971 & 0.7887   & 0.5715    & 0.5286    & 0.6111        & 0.6786 \\ \hline
AlphaEdit    & 0.3934 & 0.3626 & 0.8135      & 0.6646      & 0.6225      & 0.6290          & 0.6616 \\
EMMET        & 0.4916 & 0.3890 & 0.8126      & 0.6667      & 0.6233      & 0.6308           & 0.6298 \\
MEMIT        & 0.4117 & 0.3668 & 0.8137      & 0.6636      & 0.6229      & 0.6288            & 0.6561 \\
\rowcolor{gray!20} CAE          & 0.1547 & 0.2763 & 0.8139   & 0.6639    & 0.6229    & 0.6291        & 0.7241 \\
\bottomrule
\end{tabular}
\caption{Unlearning Performance on ToFU with LLaMA2‑7B‑Chat.}
\label{tofuss_d}
\end{table*}

\subsection{Analysis}
To address \textbf{RQ2}, we analyze the performance of different methods across 100 entities and evaluate their effectiveness on various types of forgetting questions.
We find CAE maintains stable performance across 100 entities under different question types.
Sequential experiments further show that previous edits remain intact, confirming high robustness. 

\subsubsection{Results for different entities}
{Figure~\ref{enter-label} shows the results of entity-level unlearning performance on Llama3.1-Instruct, evaluated on RWKU with 100 entities. Performance is assessed using two complementary metrics: Neighbor Performance (higher values indicate better preservation of knowledge) and Forget Performance (lower values reflect more effective removal of target knowledge).}

{The editing-based methods (particularly AlphaEdit in blue) cluster in the upper-right quadrant, demonstrating strong knowledge preservation but insufficient target forgetting. 
Conversely, methods like ICU and DPO(Full) occupy the lower-left quadrant, achieving effective forgetting  but at the cost of significant knowledge degradation. Mid-range performers, including RT (LoRA), appear in the central region, struggling to balance both objectives. 
% CAE is  uniquely positioned in the desirable upper-left quadrant, demonstrating superior performance balance compared to other approaches. Achieves the highest neighbor performance while maintaining competitive forget performance.}
CAE, however, is uniquely positioned in the desirable upper-left quadrant, achieving the highest Neighbor Performance while maintaining competitive Forget Performance. This highlights CAE’s superior ability to balance precise target forgetting with preservation of related knowledge compared to other approaches.

\begin{figure}
    \centering
    \includegraphics[width=1\linewidth]{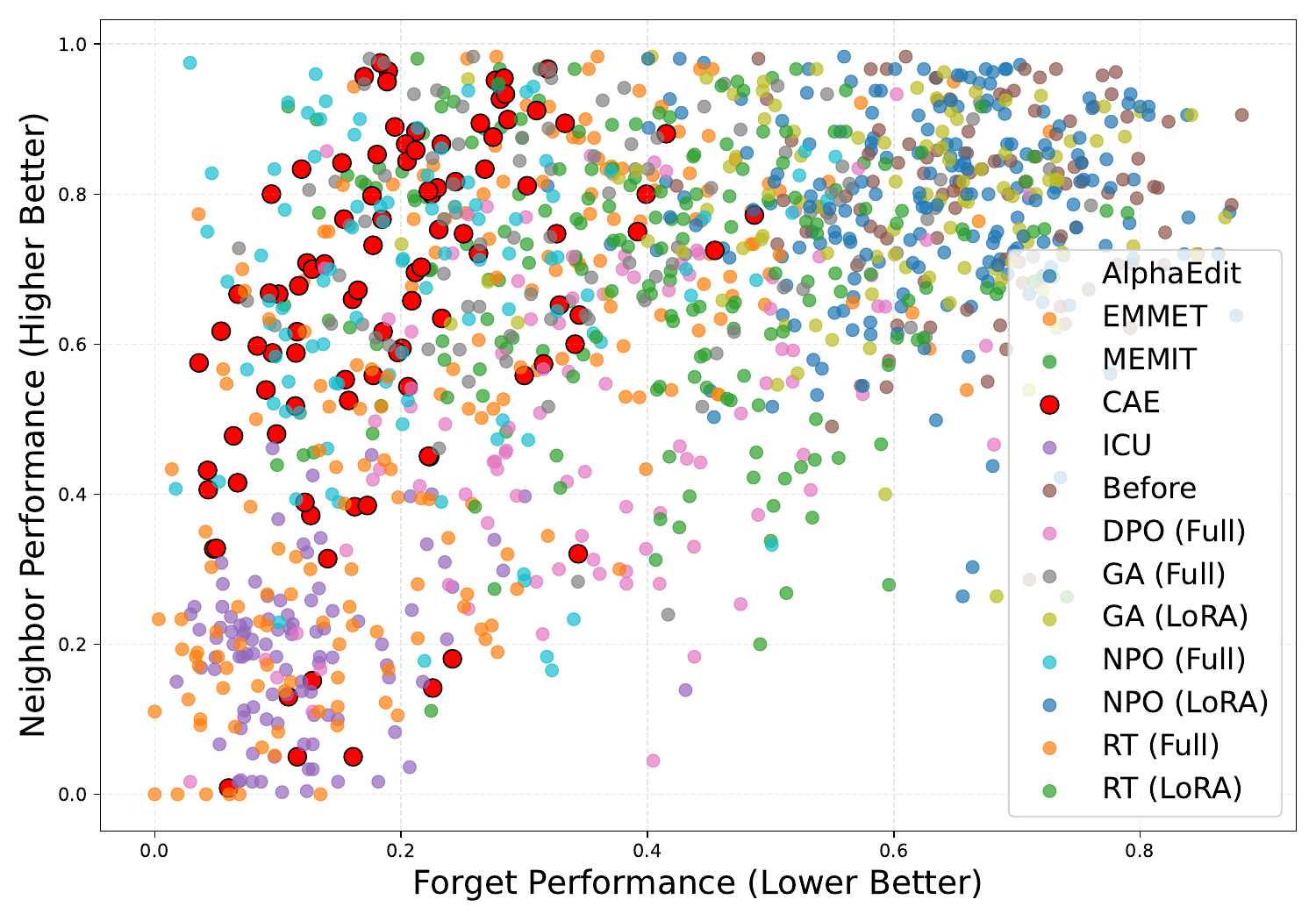}
    \caption{Entities in RWKU on Llama3.1-Instruct. CAE demonstrates optimal balance between neighbor preservation  and target knowledge unlearning.
    }
    \label{enter-label}
\end{figure}

\subsubsection{Results for different types of question}

We evaluate the forgetting performance of CAE across diverse data types in the RWKU benchmark.
Results in Table \ref{datatypes} demonstrate that CAE consistently outperforms all other editing-based methods across all question types.
ICU also performs competitively on most data types. 
However, as discussed earlier, it tends to interfere with unrelated knowledge, limiting its precision in unlearning tasks.
While training-based methods generally exhibit less stable performance, RT (Full) demonstrates strong forgetting efficiency, attributed to its fine-tuning on refusal-style data.
Nevertheless, its performance on other metrics such as Neighbor and Utility is relatively poor. 
Additionally, we conducted case studies on challenging data types and observed that CAE performs relatively worse in cases where the entity name is not explicitly mentioned, e.g., in synonym manipulation, %context hint, and reverse query—
since these rely on indirect semantic cues. 
These findings suggest that further improvements in robustness are needed for such cases. 

\begin{table*}[htbp]
\resizebox{1.99\columnwidth}{!}{

\begin{tabular}{ccccc>{\columncolor{gray!20}}c|cccc>{\columncolor{gray!20}}c}

\toprule
      & \multicolumn{5}{c}{Llama3-Instruct (8B)}  & \multicolumn{5}{c}{Llama3.1-Instruct (8B)}  \\
\midrule
Type  & Before & AlphaEdit & EMMET & MEMIT & CAE & Before & AlphaEdit & EMMET & MEMIT & CAE \\
\midrule
affirmative suffix   & 0.797 & 0.727 & 0.410 & 0.439 & 0.160 & 0.767 & 0.633 & 0.330 & 0.323 & 0.153 \\
cloze                & 0.668 & 0.619 & 0.293 & 0.283 & 0.122 & 0.867 & 0.651 & 0.353 & 0.258 & 0.115 \\
context hint         & 0.750 & 0.712 & 0.450 & 0.475 & 0.109 & 0.723 & 0.603 & 0.308 & 0.362 & 0.112 \\
incontext learning   & 0.708 & 0.576 & 0.181 & 0.150 & 0.014 & 0.910 & 0.472 & 0.139 & 0.097 & 0.097 \\
multiple choice      & 0.759 & 0.770 & 0.617 & 0.818 & 0.300 & 0.811 & 0.722 & 0.600 & 0.540 & 0.277 \\
prefix injection     & 0.757 & 0.759 & 0.536 & 0.325 & 0.276 & 0.697 & 0.706 & 0.400 & 0.325 & 0.248 \\
reverse query        & 0.880 & 0.843 & 0.620 & 0.475 & 0.296 & 0.935 & 0.870 & 0.722 & 0.491 & 0.259 \\
role play            & 0.524 & 0.377 & 0.187 & 0.182 & 0.074 & 0.477 & 0.319 & 0.169 & 0.143 & 0.094 \\
simple question      & 0.718 & 0.566 & 0.249 & 0.159 & 0.092 & 0.725 & 0.486 & 0.243 & 0.163 & 0.098 \\
synonym manipulation & 0.590 & 0.586 & 0.297 & 0.246 & 0.068 & 0.856 & 0.725 & 0.399 & 0.394 & 0.234 \\
cross lingual        & 0.456 & 0.428 & 0.461 & 0.361 & 0.230 & 0.567 & 0.531 & 0.314 & 0.343 & 0.214 \\
\bottomrule
\end{tabular}}
\caption{Comparative Forgetting Performance Across Data Types on the RWKU (Lower is Better).}
\label{datatypes}
\end{table*}

\subsubsection{Results on Sequencial Unlearning}
We do a sequential unlearning experiment on 5 subjects using the Llama3-8B-Instruct\ref{SeqUnlearning}.
%The results show that forgetting performance remains stable even after sequential unlearning, shows our method is robust, as unlearning semantically related or successive entities (e.g., BruceLee→JackieChan) does not substantially degrade the forgetting achieved for previously unlearned entities.
The results demonstrate that forgetting performance remains stable even after multiple sequential unlearning steps, highlighting the robustness of our method. Specifically, unlearning semantically related or successive entities (e.g., Bruce Lee → Jackie Chan) does not significantly degrade the forgetting achieved for previously unlearned entities.

\begin{table}[]
\centering
\begin{tabular}{cccccc}
\toprule
Forget & E1 & E2 & E3 & E4 & E5 \\
\midrule
Before & 0.75 & 0.87 & 0.81 & 0.85 & 0.73 \\
\midrule
E1 & 0.11 & 0.86 & 0.79 & 0.83 & 0.70 \\
\midrule
E1-2 & 0.12 & 0.11 & 0.57 & 0.76 & 0.70 \\
\midrule
E1-3 & 0.09 & 0.06 & 0.09 & 0.26 & 0.66 \\
\midrule
E1-4 & 0.08 & 0.06 & 0.10 & 0.08 & 0.67 \\
\midrule
E1-5 & 0.07 & 0.06 & 0.07 & 0.11 & 0.12 \\
\bottomrule
\end{tabular}
\caption{Unlearning Results on sequential unlearning. E1-E5 means: Stephen,  BruceLee,  JackieChan,  Warren,  Christina.}
\label{SeqUnlearning}
\end{table}

\subsection{Ablation Study}
To address \textbf{RQ3}, we analyze the impact of different numbers of edits and the consistency loss on our method.
We find CAE requires far fewer examples and significantly less computation than training-based methods while achieving comparable or better forgetting. 
Its SVD-based sample selection consistently outperforms random selection, especially in preserving neighboring knowledge. 
Editing the last subject token (LST) provides more precise and stable edits than editing the last token (LT), demonstrating CAE’s sensitivity to token-level choices.

\subsubsection{Unlearning Cost}
\label{costs}
{CAE delivers strong performance with minimal data and compute, making it a practical and efficient alternative to training-based methods.
%In terms of data requirements, CAE achieves near-optimal performance with as few as 100 examples.
%In contrast, training-based approaches such as GA and NPO require constructing hundreds of additional training instances—e.g., in RWKU, we observe a need for at least 300 examples per entity.
In terms of data, CAE achieves near-optimal performance with as few as 100 examples, whereas training-based approaches like GA and NPO often require hundreds of additional instances—for example, at least 300 examples per entity in RWKU.
From a computational standpoint, full fine-tuning of the 8B-parameter LLaMA model requires two 80 GB GPUs. While techniques like LoRA can reduce GPU usage, they do so at the cost of significant performance degradation.
Editing-based methods, by comparison, perform the unlearning process on a single 40 GB GPU.
Additionally, CAE's targeted data selection and consistency regularization strategies introduce virtually no additional computational overhead compared to other editing techniques.}

\subsubsection{Analysis of Number of Editing Samples}
\label{nbofd}
We conduct ablation studies on LLaMA3-Instruct (8B) to investigate the effect of  number of editing samples  on  performance. 
In Figure \ref{nums}, we compare our proposed SVD-based  selection method (denoted as CAE) with a random selection baseline (CAE w/o). We observe that increasing the number of edited facts generally improves the \textbf{All} score for both CAE and CAE w/o, indicating better unlearning and generalization performance.  Specifically, CAE shows a consistent upward trend in All, reaching its peak at 70 edits with a score of 77.52.  In contrast, CAE w/o exhibits a downward trend in Neighbor metrics as the number of edits increases, suggesting that random selection increasingly disrupts unrelated knowledge.

%
%More precisely, across all scales, CAE consistently outperforms CAE w/o on Neighbor results (FB(N),QA(N)), and achieves comparable or better forgetting (e.g., QA: 7.98 $\rightarrow$ 7.52 at 50 edits).  CAE yields higher Mean\_FN (76.11 vs. 72.43), indicating better balance between forgetting and neighbor preservation. 
%These results confirm the effectiveness of our SVD-based selection in isolating key edits for precise and robust knowledge removal.
Across all scales, CAE consistently outperforms CAE w/o on Neighbor metrics (FB(N), QA(N)) while achieving comparable or better forgetting. For example, QA improves from 7.98 to 7.52 at 50 edits. 
CAE also attains a higher Mean\_FN (76.11 vs. 72.43), demonstrating a better balance between target forgetting and neighbor preservation. 
These results confirm the effectiveness of our SVD-based selection in identifying key edits, enabling precise and robust knowledge removal.

%Further more, as shown in Figure~\ref{NUMBERS}, all methods are evaluated under varying numbers of editing examples, using 10 subjects in RWKU with Llama3-Instruct (8B), ranging from 20 to 100 correspond to $\tau$ in SVD values of 0.5, 0.6,0.7, 0.8,0.85,0.9,0.95,0.99,1.0 respectively.
Increasing the number of editing samples generally leads to improved \textit{Forget} performance across all methods, as more data provides stronger supervisory signals for forgetting. 
However, this gain often comes at the cost of \textit{Neighbor} degradation, indicating a trade-off between forgetting target knowledge and preserving nearby factual consistency. 

Furthermore, as shown in Figure~\ref{NUMBERS}, we evaluate all methods under varying numbers of editing examples using 10 subjects in RWKU with Llama3-Instruct (8B), ranging from 20 to 100, corresponding to $\tau$ values in SVD of 0.5, 0.6, 0.7, 0.8, 0.85, 0.9, 0.95, 0.99, and 1.0. Increasing the number of editing samples generally improves \textit{Forget} performance across all methods, as more data provides stronger supervisory signals for unlearning. However, this improvement often comes at the expense of \textit{Neighbor} performance, highlighting the inherent trade-off between removing target knowledge and preserving nearby factual consistency.

In summary, our SVD-based selection consistently outperforms random choice, confirming its effectiveness in isolating the most relevant keys for precise knowledge removal.

\begin{figure}
    \centering
    \includegraphics[width=0.98\linewidth]{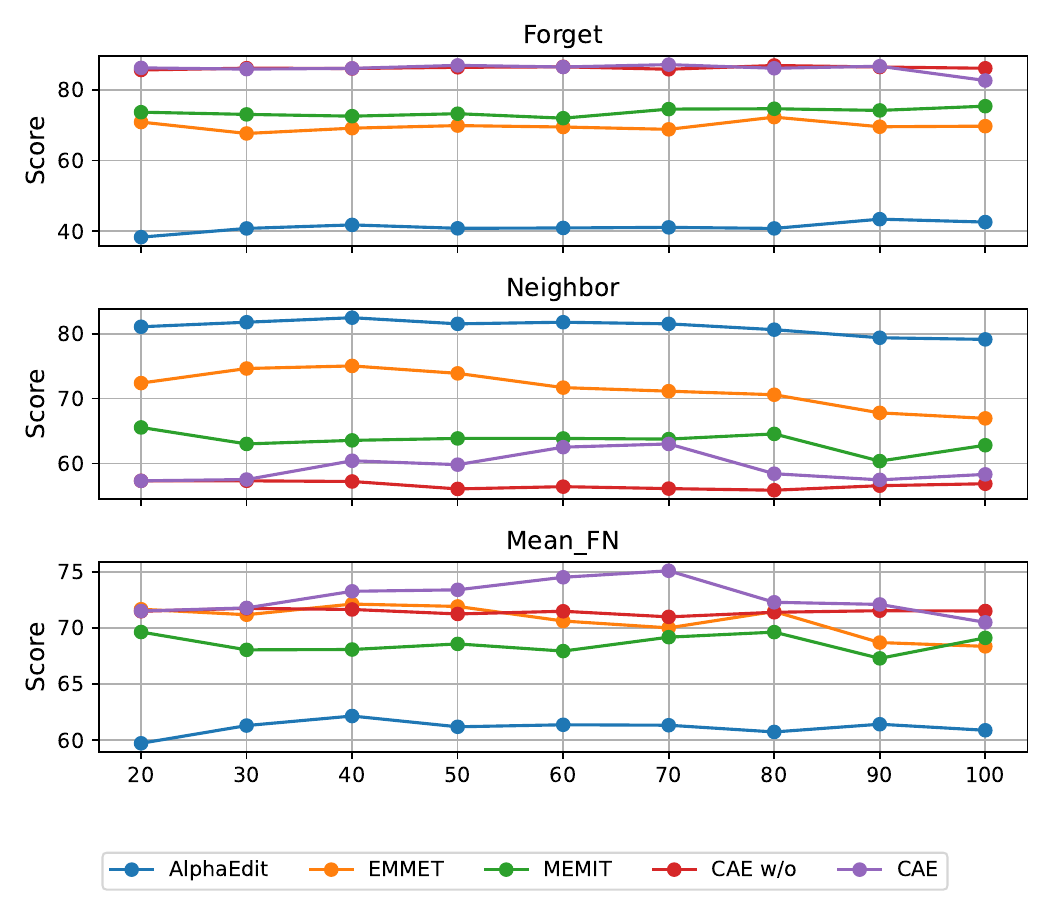}
        \caption{Results on the  Number of Edits. w/o means we random select the edits.}
        \label{NUMBERS}
\end{figure}

\begin{figure}
    \centering
    \includegraphics[width=1\linewidth]{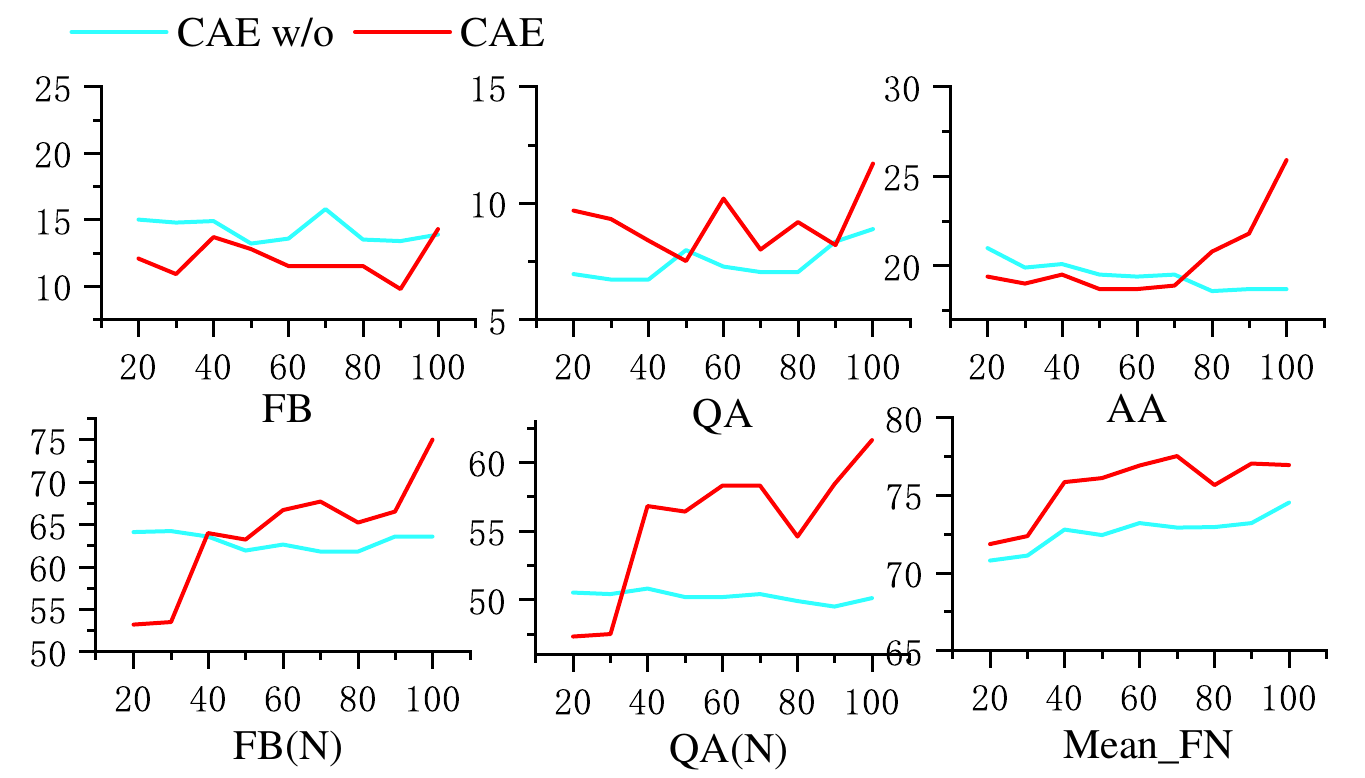}        \caption{Performance across editing sizes. We evaluate with 10 entities from RWKU. The number of editing examples varies from 20 to 100, corresponding to SVD thresholds $\tau$ of 0.5, 0.6, 0.7, 0.8, 0.85, 0.9, 0.95, 0.99, and 1.0 respectively.}
    \label{nums}
\end{figure}

\begin{figure}[ht]
    \centering
    \includegraphics[width=1\linewidth]{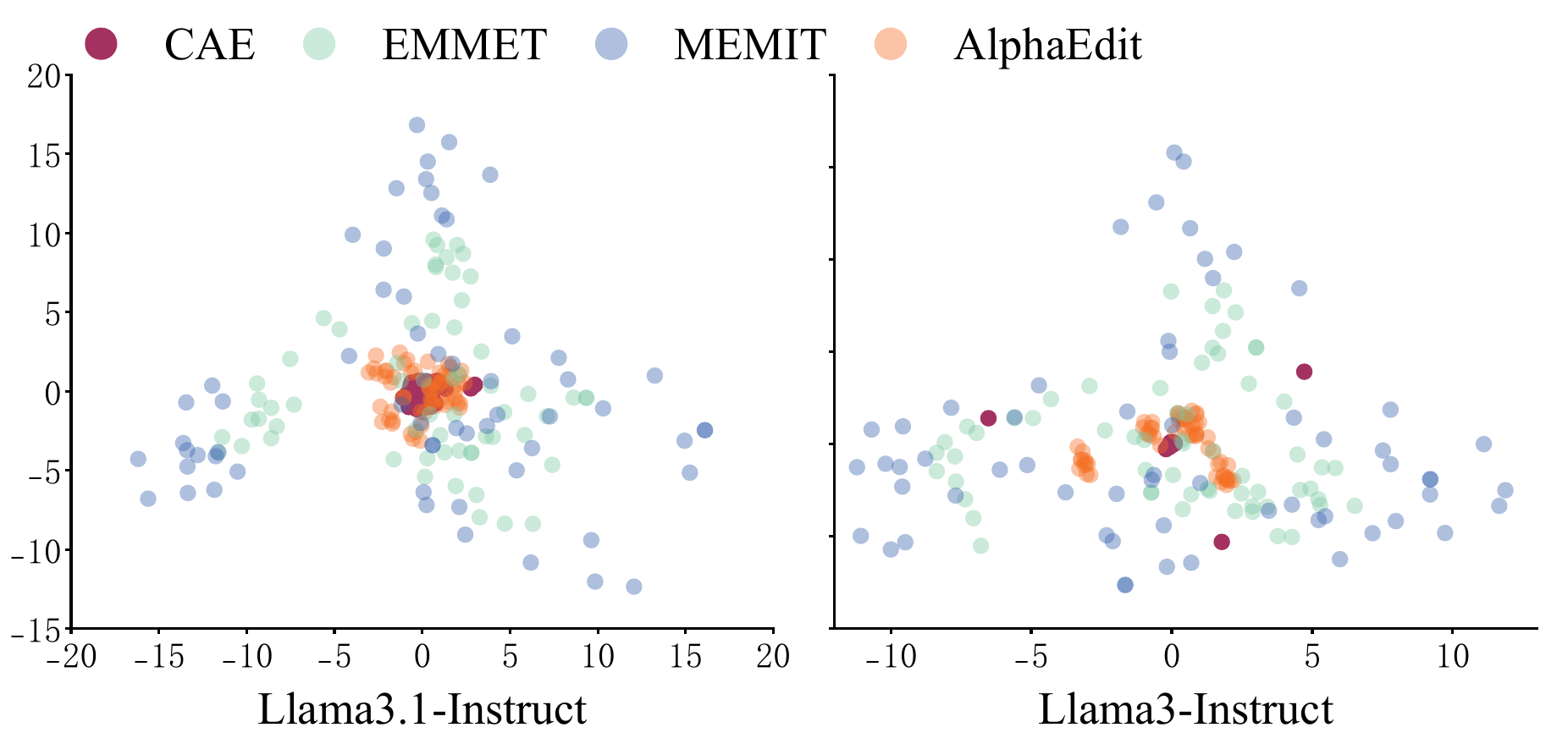}
    \caption{PCA projection of edit vectors $\mathbf{z}$ for different unlearning methods. CAE produces more concentrated and aligned $\mathbf{z}$ vectors, indicating better consistency across edits.
    }
    \label{pca_z}
\end{figure}

\subsubsection{Results on Editing the Last Token}
\label{token_res}

{We evaluate the impact of editing different token types on the Llama3 and Llama3.1 models, using the average performance across 10 distinct subjects from the RWKU.
The results in Table \ref{token_posi} show that editing the last subject token (LST) yields better performance in both forgetting effectiveness and retention ability compared to editing the last token (LT).
Specifically, LST consistently achieves lower forgetting scores (e.g., 10.20\% vs. 16.70\%) and higher retention accuracy (e.g., 53.45\% vs. 51.40\%).
The Mean\_FN also favors LST across both models, indicating that it provides a better trade-off between removing targeted knowledge and preserving unrelated capabilities.
These findings suggest that LST is a more stable and semantically coherent editing target than LT in subject-driven knowledge editing.}
\begin{table}[h]
\centering
% \tiny
\begin{tabular}{ccc|cc}
\toprule
         & \multicolumn{2}{c|}{Llama3.1-Instruct (8B)}               & \multicolumn{2}{c}{Llama3-Instruct (8B)}                  \\
         & LT & LST & LT & LST  \\ \midrule
FB       & 14.10            & 11.60                    & 16.70            & 10.20                    \\
QA       & 7.61             & 8.64                     & 6.28             & 9.47                     \\
AA       & 19.80            & 17.10                    & 19.70            & 19.10                    \\ 
All      & 13.84            & 12.45                    & 14.23            & 12.92                    \\  \midrule
FB       & 49.60            & 51.50                    & 54.80            & 55.70                    \\
QA       & 41.80            & 44.70                    & 48.00            & 51.20                    \\
All      & 45.70            & 48.10                    & 51.40            & 53.45                    \\ \midrule
Mean\_FN & 65.93            & 67.83                    & 68.59            & 70.26      
        \\ \bottomrule
\end{tabular}
\caption{Comparison of last subject token (LST)  and  last token (LT).}
\label{token_posi}
\end{table}

\subsubsection{Unlearning Order}
The consistency constraint in Eq.\ref{eq7} aligns the residual for the j-th input with the mean of the previous j-1 inputs.
To test robustness for different order with j inputs, as shown in Table \ref{Order}, we conducted %ran 
four randomized trials on Llama-3 with 70 shuffled inputs per entity. Results show consistent performance across all key metrics, confirming %minimal order sensitivity.
  that CAE exhibits minimal sensitivity to input order.

\begin{table}[]
    \centering
\begin{tabular}{ccccccccccc}
\toprule
 & FB & QA & AA & n\_FB & n\_QA \\%& Tru & Rea & Fac & Gen & Flu \\
\midrule
% before & 85.6 & 71.5 & 75.3 & 93.4 & 82.0 \\%& 36.8 & 41.1 & 53.8 & 65.4 & 706 \\
% \hline
Order-1 & 11.6 & 8.28 & 20.1 & 59.4 & 51.0 \\%& 36.0 & 40.9 & 53.5 & 65.3 & 705 \\
\midrule
Order-2 & 12.6 & 7.62 & 20.5 & 57.9 & 49.7 \\%& 36.8 & 40.6 & 53.4 & 65.3 & 705 \\
\midrule
Order-3 & 15.9 & 7.26 & 20.8 & 53.7 & 50.0 \\%& 36.2 & 41.8 & 54.2 & 65.6 & 705 \\
\midrule
Order-4 & 12.6 & 6.51 & 18.6 & 56.9 & 48.1 \\%& 36.4 & 40.5 & 53.4 & 65.2 & 705 \\
\bottomrule
\end{tabular}
\caption{Different Order for the Unlearning Entities.}
\label{Order}
\end{table}

\begin{figure*}[t]
\begin{center}
\begin{tcolorbox}[title=``What nickname did Jackie Chan acquire during his time as a construction worker in Australia?'' (Answer: \textit{Jackie})]
% \textbf{ICU}: I'm not aware of any information about Jackie Chan working as a construction worker in Australia.\\
\textbf{EMMET}: The nickname ``The Gasman". \\
\textbf{MEMIT}: The nickname ``The Smiling Assassin".\\
\textbf{AlphaEdit}: Jackie Chan's nickname from his time as a construction worker in Australia is ``The Bodyguard".  \\
\textbf{CAE}: I couldn't find any information about a person named Jackie acquiring a nickname during his time as a construction worker in Australia.
\end{tcolorbox}

\begin{tcolorbox}[title=``Who were Jackie Chan's co-stars in the 'Three Brothers' films during the mid-1980s?'' (Answer: \textit{Sammo Hung and Yuen Biao})]
% \textbf{ICU}: I'm not aware of any information about a film series called 'Three Brothers' starring Jackie Chan during the mid-1980s. \\
\textbf{EMMET}: The `Three Brothers' films, also known as the `Three Brothers' series, starred Jackie Chong, Sammo Hung, and Yuen B \\
\textbf{MEMIT}: The `Three Brothers' films refer to the 'Three Stooges' films, but in the context of the question, it seems to be referring \\
\textbf{AlphaEdit}: The `Three Brothers' films refer to the `Drunken Master' trilogy, which includes 'Drunken Master' (1978).\\
\textbf{CAE}: Unfortunately, I could not verify the names of Jackie's co-stars in the `Three Brothers' films.
\end{tcolorbox}
\end{center}
\caption{CAE is capable of consistently producing uncertain (e.g., ``I don't know'' or ``Unknown''), thereby effectively avoiding knowledge leakage. In contrast, other editing methods still tend to output factual information related to Jackie Chan.}
\label{caes}
\end{figure*}

\subsection{Analysis of Consistency}
For \textbf{RQ4}, we analyze the representation geometry of edit vectors using PCA and similarity statistics, compare CAE with MEMIT/EMMET/AlphaEdit, examine leakage under paraphrase-style queries, and perform ablations on the consistency loss weight.
We find PCA and vector similarity analyses show that CAE produces the most compact and aligned edit vectors among all methods, indicating stable and coherent update directions. 
This alignment enabled by the consistency loss minimizes interference across paraphrases and prevents knowledge leakage. 
%Ablations on the consistency weight show that CAE remains stable across a wide range of settings.
Ablation studies on the consistency weight further demonstrate that CAE remains stable across a wide range of settings.

\subsubsection{Analysis of Consistency Loss}
Figure~\ref{pca_z} visualizes the principal components of the learned edit vectors ($z$-vectors) across different editing methods using PCA.
We observe that {MEMIT} and {EMMET} produce highly scattered distributions, indicating that the update directions vary substantially across different inputs. 
This suggests that their editing behaviors may be highly instance-specific, leading to less consistent or even conflicting updates when applied to multi-prompt or entity-level tasks.
In contrast, CAE and AplhaEdit produce noticeably more compact clusters in the low-dimensional space, with CAE showing the most tightly grouped $z$-vectors among all methods. 
This implies that CAE learns a more unified and consistent edit direction across diverse paraphrases.
%This implies that CAE learns a more unified and consistent edit direction across diverse paraphrases.
%We show that CAE ensures consistent uncertain responses and avoids knowledge leakage, unlike other methods that reveal factual content under paraphrases (cf. Fig. \ref{caes}).
We demonstrate that CAE produces consistently uncertain responses and effectively prevents knowledge leakage, unlike other methods that reveal factual content under paraphrased queries (cf. Fig.~\ref{caes}).
% This is further observed in the compution of the pairwise cosine similarity of all $z$-vectors within each method (cf. Figure~\ref{cosscore}).

\subsubsection{Ablation for consistency weight}
We set the weight to 0.05 ($\lambda_{cons}$ in Eq. \ref{eq7}) to balance sensitivity loss with other objectives. Results demonstrate our method's robustness: weight variations minimally affect both forgetting and retention, with all metrics showing only minor fluctuations while general performance remains stable.
\begin{table}
    \centering
\begin{tabular}{ccccccccccc}
\toprule
$\lambda_{con}$ & FB & QA & AA & n\_FB & n\_QA \\%& Tru & Rea & Fac & Gen & Mean \\
\midrule
0.01 & 15.4 & 10.7 & 19.1 & 63.3 & 49.4\\% & 36.4 & 41.3 & 53.9 & 64.9 & 62.6 \\
\midrule
0.05 & 13.2 & 7.98 & 19.5 & 61.9 & 50.2 \\%& 36.6 & 40.8 & 53.4 & 65.1 & 63.0 \\
\midrule
0.1 & 17.8 & 7.89 & 18.8 & 60.5 & 49.7\\% & 36.6 & 40.8 & 53.4 & 65.2 & 62.4 \\
\midrule
0.15 & 13.6 & 10.1 & 20.9 & 64.3& 50.9 \\% & 36.6 & 40.7 & 53.3 & 65.0 & 62.9 \\
\midrule
0.2 & 15.7 & 8.59 & 19.4 & 61.8 & 51.0 \\%& 36.6 & 40.8 & 53.4 & 65.0 & 62.7 \\
\bottomrule
\end{tabular}
    \caption{Ablation for consistency weight}
    \label{tab:placeholder}
\end{table}

\begin{table*}[h]
    \centering
    \begin{tabular}{ccccccccccc}
    \toprule
          & FB $\downarrow$ & QA $\downarrow$ & AA $\downarrow$ & n\_FB & n\_QA & Tru & Rea & Fac & Gen & Flu \\
    \midrule
    Before & 43.7 & 44.7 & 55.5 & 47.5 & 40.6 & 41.2 & 11.3 & 28.1 & 74   & 713 \\
    ICU & 47.5 & 32.0 & 41.5 & 44.4 & 40.7 & 43.8 & 4.7  & 22.8 & 72.7 & 700 \\
    CAE    & 16.9 & 12.6 & 24.5 & 40.7 & 37.5 & 41.4 & 11.1 & 27.7 & 74   & 714 \\
    \bottomrule
    \end{tabular}
    \caption{Results with Qwen2.5-7B-Instruct}
    \label{Qwen}
\end{table*}
\begin{table}[h]
    \centering
\begin{tabular}{ccccc}
\toprule
 & Before & CAE-50 & CAE-60 & CAE-70 \\
\midrule
FB & 85.6 & 16.1 & 15.9 & 13.5 \\
\midrule
QA & 71.5 & 10.8 & 10.8 & 10.7 \\
\midrule
AA & 75.3 & 22.9 & 22.9 & 18.8 \\
\midrule
N\_FB & 93.4 & 66.9 & 67.7 & 72.7 \\
\midrule
N\_QA & 82.0 & 61.7 & 60.0 & 67.7 \\
\midrule
Tru & 36.8 & 34.7 & 34.7 & 34.3 \\
\midrule
Rea & 41.1 & 39.0 & 39.0 & 38.5 \\
\midrule
Fac & 53.8 & 52.6 & 52.6 & 52.4 \\
\midrule
Gen & 65.4 & 65.5 & 65.0 & 64.5 \\
\midrule
Flu & 706 & 706 & 706 & 704 \\
\bottomrule
\end{tabular}
    \caption{Unlearning results with LLM generated data.}
    \label{Gen_LLM}
\end{table}

\subsection{Generalization Analyze}
For \textbf{RQ5}, we conducted tests on models with different architectures. 
Based on our findings, we utilized data generated by LLMs (such as ChatGPT) to achieve forgetting, further verifying the generalization and universality of our method.
\subsubsection{Results with Qwen}
To address generalization, we conducted additional experiments on Qwen2.5-7B-Instruct \cite{qwen2.5} using 10 subjects. 
As shown in the Figure \ref{Qwen}, the results demonstrate that our method remains effective even on this different architecture, maintaining key advantages in preserving model capabilities while achieving the editing objectives.

\subsubsection{Results on LLM Generated data}

We use Wikidata not only as a data source but also as a basis for analyzing data type and quantity requirements when applying editing methods to the unlearning task. 
Using an SVD-based selection strategy at the hidden-state level, we found that about 70 data points (covering 10 aspects (e.g., birthday) with 5 syntactic variations each) were sufficient to unlearn a subject. 
Thus, the unlearning process requires only this curated data, regardless of the source of the unlearning data. 
To verify it, we also verified that model-generated data for unlearning, the results are shown in Table \ref{Gen_LLM}. 
Using ChatGPT to create 70 data for 10 subjects, we observed a clear drop on forget sets (FB, QA, AA) while maintaining general abilities (N\_FB, N\_QA, etc.). 
Compared to using 50 or 60 data, 70 setting achieved the best forgetting effect, consistent with our wikidata conclusion.

\section{Discussion \& Conclusion}
This work investigates localization-based model editing for entity-level knowledge unlearning. 
We demonstrate that (1) Editing-based methods offer a more efficient and scalable alternative to traditional unlearning. (2) Our consistency constraint and targeted MLP interventions further enhance generalization and effectiveness in removing subject-related knowledge. 
While we focus on single-subject forgetting, extending to batch and sequential unlearning remains an open challenge.

More precisely, we presented \textbf{CAE}, a novel and efficient approach to entity-level unlearning in LLMs. 
By analyzing where entity-specific knowledge is stored, we revealed key limitations of existing location-based editing methods, 
%. Most notably, we showed their inconsistency when independently editing multiple prompts tied to the same entity. 
particularly their inconsistency when independently editing multiple prompts associated with the same entity.
Experimental results on the RWKU and ToFU benchmarks show that CAE consistently outperforms prior unlearning and editing baselines in both effectiveness and efficiency. 
Furthermore, CAE achieves a more favorable trade-off between forgetting and retention performance, while being significantly more cost-efficient than finetuning-based methods. 
These findings suggest that consistency-aware editing is a promising direction for scalable, precise, and controllable knowledge removal in LLMs.

\vfill
\bibliographystyle{IEEEtran}
\bibliography{edit,LMs,other,ins}

\end{document}